\pdfoutput=1

\documentclass[11pt]{article}
\usepackage{subcaption}

\usepackage[hyperref]{emnlp2021}
\usepackage{times}
\usepackage{latexsym}

\usepackage[T1]{fontenc}

\usepackage[utf8]{inputenc}

\newenvironment{citemize}{\begin{list}{$\bullet$}{\topsep=.2\smallskipamount\itemsep=0pt\parsep=1pt\labelwidth=.5em}}{\end{list}}
\newenvironment{cenumerate}{\begin{list}{\labelenumi}{\usecounter{enumi}\topsep=.2\smallskipamount\itemsep=0pt\parsep=1pt\labelwidth=.5em}}{\end{list}}

\usepackage{multirow}
\usepackage{booktabs}

\usepackage[colorinlistoftodos,textwidth=12em]{todonotes}
\reversemarginpar

\usepackage{microtype}

\newcommand{\titletext}{Understanding Model Robustness to User-generated Noisy Texts}
\title{\titletext}

\def\MLNoise{\mbox{KaziText}}

\author{
  Jakub Náplava \and Martin Popel \and Milan Straka \and Jana Straková\\
  Charles University, \\
  Faculty of Mathematics and Physics, \\
  Institute of Formal and Applied Linguistics \\
  \texttt{\{naplava,popel,straka,strakova\}@ufal.mff.cuni.cz}
}

\date{}

\newcommand{\beginsupplement}{%
  \setcounter{table}{0}%
  \renewcommand{\thetable}{S\arabic{table}}%
  \setcounter{figure}{0}%
  \renewcommand{\thefigure}{S\arabic{figure}}%
  \setcounter{section}{0}%
  \renewcommand\thesection{S\arabic{section}}%
}

\makeatletter
\newcommand{\titlecopy}{\begingroup\def\thefootnote{\fnsymbol{footnote}}\def\@makefnmark{\hbox to 0pt{$^{\@thefnmark}$\hss}}\twocolumn[\@maketitle] \@thanks\endgroup\setcounter{footnote}{0}}
\makeatother

\usepackage[absolute]{textpos}
\begin{document}
\begin{textblock}{16}(0,0.1)\centerline{This paper was published in \textbf{W-NUT 2021} -- please cite the published version {\small\url{https://aclanthology.org/2021.wnut-1.38}}.}\end{textblock}
\titlecopy
\begin{abstract}
Sensitivity of deep-neural models to input noise is known to be a challenging problem. In NLP, model performance often deteriorates with naturally occurring noise, such as spelling errors. To mitigate this issue, models may leverage artificially noised data. However, the amount and type of generated noise has so far been determined arbitrarily. We therefore propose to model the errors statistically from grammatical-error-correction corpora. We present a thorough evaluation of several state-of-the-art NLP systems' robustness in multiple languages, with tasks including morpho-syntactic analysis, named entity recognition, neural machine translation, a subset of the GLUE benchmark and reading comprehension. We also compare two approaches to address the performance drop: a) training the NLP models with noised data generated by our framework; and b) reducing the input noise with external system for natural language correction. The code is released at \url{https://github.com/ufal/kazitext}.
\end{abstract}

\section{Introduction}

Although there has recently been an amazing progress in variety of NLP tasks~\citep{vaswani2017attention, devlin2018bert} with some models even reaching performance comparable to humans on certain domains~\citep{ge2018reaching, popel2020transforming}, it has been shown that the models are very sensitive to noise in data~\citep{belinkov2017synthetic, rychalska2019models}.


Multiple areas of NLP have been studied to evaluate the effect of noise in data \cite{belinkov2017synthetic,heigold2018robust, ribeiro2018semantically,glockner2018breaking} and a framework for text corruption to test NLP models robustness is also available \cite{rychalska2019models}. However, all these systems introduce noise in a custom-defined, arbitrary level and typically for a single language.



\eject
We suggest modeling natural noise statistically from corpora and we propose a framework with the following distinctive features:

\begin{itemize}
    \item The error probabilities are estimated on real-world grammatical-error-correction corpora.
    \item The intended noisiness can be scaled to a desired level and
    various aspects (types) of errors can be turned on/off to test the NLP systems robustness to specific error types.
\end{itemize}

Furthermore, we also present a thorough evaluation of several current state-of-the-art NLP systems' with varying level of data noisiness and a selection of error aspects in multiple languages. The NLP tasks include morpho-syntactic analysis, named entity recognition, neural machine translation, a subset of GLUE benchmark and reading comprehension. We conclude that:

\begin{itemize}
    \item The amount of noise is far more important than the distribution of error types.
    \item Sensitivity to noise differs greatly among NLP tasks. While tasks such as lemmatization require correcting the input text,
    only an~approximate understanding is sufficient for others.
\end{itemize}

We also compare two approaches for increasing models robustness to noise: training with noise and external grammatical-error-correction (GEC) preprocessing. Our findings suggest that training with noise is beneficial for models with large capacity and large training data (neural machine translation), while the preprocessing with grammatical-error-correction is more suitable for limited-data classification tasks, such as morpho-syntactic analysis.

Finally, we also offer an evaluation on authentic noise: We assembled a new dataset with authentic Czech noisy sentences translated into English and we evaluate the noise-mitigating strategies in the neural machine translation task on this dataset.




\section{Related Work}

Many empirical findings have shown the fact that data with natural noise deteriorate NLP systems performance. \newcite{belinkov2017synthetic} found that natural noise such as misspellings and typos cause significant drops in BLEU scores of character-level machine translation models.
To increase the model's robustness, they trained the model on a mixture of original and noisy input and found out that it learnt to address certain amount of errors. Similar findings were observed by \newcite{heigold2018robust} who tested machine translation and morphological tagging 
under three types of word-level errors.

\newcite{ribeiro2018semantically} defined a set of substitution rules that produce semantically equivalent text variants. They used them to test systems in machine comprehension, visual question answering and sentiment analysis. \newcite{glockner2018breaking} created a new test set for natural language inference and showed that current systems do not generalize well even for a single-word replacements by synonyms and antonyms.

\newcite{rychalska2019models} implemented a framework for introducing multiple noise types into text such as removing or swapping articles, rewriting digit numbers into words or introducing errors in spelling. They found out in four NLP tasks that even recent state-of-the-art systems based on contextualized word embeddings
are not completely robust against such natural noise. They also re-trained the systems on noisy data and observed improvements for certain error types.

Similarly to \newcite{rychalska2019models}, we also developed a general framework that allows to test a variety of NLP tasks. The difference is that we estimate the probabilities of individual error types from real-world error corpora. This makes the generated sentences more similar to what humans would do.
Moreover, since we defined the individual error types with no language-specific rules, we can apply it to multiple languages with an available annotated grammatical-error corpus.

Grammatical-error corpora are typically used as training data for estimating error statistics in GEC systems. In a setting similar to ours, \citet{choe2019neural,rozovskaya2017adapting} also estimated error statistics and used them to generate additional training data for GEC systems. However, compared to our approach, they defined only a small set of predefined error categories and used it specifically for training GEC systems whereas we also use it to asses model performance in noisy scenarios.

\medskip
\noindent{\bf Authentic Noise Evaluation} The growing interest in developing production-ready machine translation
models that are robust to natural noise resulted in the First Shared Task on Machine Translation Robustness \citep{li2019findings}. The shared task used the MTNT dataset \citep{michel2018mtnt}, which consists of noisy texts collected from Reddit and their translations between English and French and English and Japanese.

\medskip
\noindent{\bf Improving Model Robustness Using Noisy Data} Majority of research on improving model robustness is dedicated to training on a mixture of original and noisy data. The same procedure is usually used for generating both the test corpus and training data \citep{belinkov2017synthetic, heigold2018robust, ribeiro2018semantically, rychalska2019models}.

To generate synthetic training data, researchers in machine translation and GEC often use so called back-translation~\citep{sennrich2015improving}. A reverse model translating in the opposite direction (i.e. from the target language to the source language or from the clean sentence into noisy sentence, respectively) is trained~\citep{rei2017artificial, naplava2017natural, kasewa2018wronging, xie2018noising}. It is then used on the corpus of clean sentences to generate noisy input data. While this approach might generate high-quality synthetic data, it requires large volumes of training data. 


We evaluate two approaches to alleviate performance drop on noisy data: We either train the system on a mixture of synthetic (generated statistically from real error corpora) and original authentic data; or we use an external grammatical-error-correction system to correct the noisy data before inputting them to the system itself. We are not aware of any other work that compares these two approaches and we believe that both approaches may be beneficial under certain conditions.

\section{Modeling Natural Noise from Corpora}

Robustness of NLP models to natural noise would ideally be evaluated on texts with authentic noise, with error corrections annotated by humans. (We present such authentic data evaluation in Section~\ref{sec:authentic}.) This perfect-world setting, however, requires an immense annotation effort, as multiple target domains have to be covered by well-educated human annotators for multiple NLP tasks in a range of languages. 
To ease the annotation burden, we propose a new framework, named \MLNoise{}, for introducing natural-like errors in a text.

The core of \MLNoise{} 
is a set of several common error type classes, \textit{aspects} (following naming convention of \citealp{rychalska2019models}).
The \textit{aspects} are composable (can be combined) and the probability of the aspect manifestation as well as the aspect's internal probabilities are estimated from grammatical-error-correction corpora.

\subsection{Noising Aspects}

\looseness1
One of the main objectives of our error aspects' design was to avoid manually designed rules, especially those derived from a single language. An~ideal approach, automatically inferring the aspects themselves, is however limited by the amount of available data. Therefore, we defined a rich set of aspects which can be estimated from the data:
\begin{cenumerate}
    \item \textbf{Diacritics} Strip diacritics either from a whole sentence or randomly from individual characters.
    \item \textbf{Casing} Change casing of a word, distinguishing between changing the first letter and other.
    \item \textbf{Spelling} Insert, remove, replace or swap individual characters (\textit{wrong}~$\rightarrow$~\textit{worng}) or use ASpell\footnote{\url{http://aspell.net/}} to transform a word to other existing word (\textit{break}~$\rightarrow$~\textit{brake}).
    \item \textbf{Suffix/Prefix} Replace common suffix (\textit{do}~$\rightarrow$~\textit{doing}) and prefix (\textit{bid}~$\rightarrow$~\textit{forbid}).
    
    \item \textbf{Punctuation} Insert, remove or replace punctuation.
    \item \textbf{Whitespace} Remove or insert spaces in text.
    \item \textbf{Word Order} Reorder several adjacent words.
    \item \textbf{Common Other} Insert, replace or substitute common phrases as seen in data (\textit{the}~$\rightarrow$~\textit{a}, \textit{a lot of}~$\rightarrow$~\textit{many}). This is the aspect which should learn language specific rules.
\end{cenumerate}

The natural errors found in real-world texts rarely fall into mutually exclusive categories. Casing errors are also spelling errors; common other aspect covers all other aspects. Therefore, some of the aspects naturally overlap. We therefore opted for evaluating the aspects in a \textit{cumulative} manner in the designed order. 


When designing the order of the aspects, our goal was to respect the natural inclusion of aspects and also error severity. We therefore start with diacritical-only changes, given that for example in Czech, users may deliberately write without diacritics. We then add casing changes, spelling errors and then suffix/prefix changes (the latter being morphologically motivated spelling errors). The first four aspects do not modify tokenization, making them suitable for tokenization-dependent tasks like POS tagging or lemmatization.

The remaining aspects change the number of tokens or token boundaries. The punctuation, whitespace and word order aspects are relatively independent, with the common other aspect covering all of them and thus being the last one.



\subsection{Estimating Noising Aspects Probabilities}

We use grammatical-error-correction (GEC) datasets to estimate probabilities of individual aspects. The GEC datasets are distributed in M2 format,\footnote{GEC file format since the CoNLL-2013 shared task} which for a tokenized input noisy sentence contains a set of correcting edits. Each correcting edit contains the corresponding input sentence span, the correction itself and the error type. The noising aspect probabilities are estimated by frequency analysis.\footnote{We refer to the published source code for details.} 
To accurately model the distribution of amount of errors in different sentences, we also measure the standard deviation of the token edit probability per sentence.

We collected M2 files from various grammatical-error-correction corpora in 4 languages: English, Czech, Russian and German. The majority of annotated content comes from Second Learners of the particular language and in addition, more speaker groups are available in English and Czech:

\begin{table*}[t]
    \centering
    \small
    \setlength{\tabcolsep}{3.3pt}
    \begin{tabular}{llrrl}
        \toprule
        \multicolumn{1}{c}{Language} & \multicolumn{1}{c}{Corpus} & \multicolumn{1}{c}{Sentences} & \multicolumn{1}{c}{Error rate} & \multicolumn{1}{c}{Domain} \\
        \midrule
        \multirow{4}{*}{English} & \textit{NUCLE}~\cite{dahlmeier2013building} & 57\,151 & 6.6\%  & SL \\ 
                                 & \textit{FCE}~\cite{yannakoudakis2011new} & 33\,236  & 11.5\%  & SL  \\ 
                                 & \textit{W\&I}~\cite{yannakoudakis2018} & 37\,704 & 11.7\%  & SL \\ 
                                 & \textit{LOCNESS}~\cite{granger1998} & 988 & 4.7\%   & native students \\ 
        \midrule
        \multirow{4}{*}{Czech} &  Romani part of \textit{AKCES-GEC}~\cite{akces_gec} & 16\,030 & 20.3\% & Romani heritage speakers \\ 
                &   SL part of \textit{AKCES-GEC}~\cite{akces_gec} & 31\,341 & 22.1\% & SL essays \\
                & \textit{Natives Informal} & 11\,608 & 15.6\% & web discussions\\
                & \textit{Natives} & 7\,696 & 5.8\% & native students\\ 
        \midrule
        German & \textit{Falko-MERLIN}~\cite{boyd2018using} & 24\,077 & 16.8\% & SL essays\\ 
        \midrule
        Russian & \textit{RULEC-GEC}~\cite{rozovskaya2019grammar} & 12\,480 & 6.4\% & SL, heritage speakers \\
          \bottomrule
    \end{tabular}
    \caption{Comparison of used GEC corpora in size, token error rate and domain. SL = second language learners.}
    \label{tab:data_overview}
\end{table*}

\begin{citemize}
    \item English  \begin{citemize}
        \item Natives: LOCNESS v2.1~\cite{granger1998}
        \item Second Learners: NUCLE~\cite{dahlmeier2013building}, FCE~\cite{yannakoudakis2011new}, Write \& Improve~\cite{yannakoudakis2018} 
    \end{citemize}
    \item Czech \begin{citemize} 
        \item Natives: essays of Czech primary schools students, in submission process
        \item Natives Informal: web discussions data, in submission process
        \item Second Learners:\\ AKCES-GEC~\cite{akces_gec}
        \item Romani: AKCES-GEC~\cite{akces_gec} -- Romani ethnic minority 
        children and teenagers
        using Czech
    \end{citemize}
    \item German (Second Learners): Falko-MERLIN GEC Corpus~\cite{boyd2018using}
    \item Russian (Second Learners):  RULEC-GEC~\cite{rozovskaya2019grammar}
\end{citemize}

\noindent An overview of the sizes and error rates of the datasets above is presented in Table~\ref{tab:data_overview}.

We call the resulting single file containing all aspect probabilities for one group of speakers a \textit{profile}. The \textit{profile} therefore describes the grammatical style of a particular given group of users, derived from M2 file annotations.  

Each profile has a development and test version originating from the respective M2 development and test files. The test profiles are used for synthesising data intended directly for assessing models' performance in noisy setting while the development profile is intended for creating data for training the models.

\subsection{Adjusting the Percentage of Token Edits}

In order to reach an intended percentage of \textit{token edits}, which directly corresponds to the amount of noise in the generated data, we correspondingly scale the aspects' probabilities. We refer to the percentage of token edits in the original corpus as a \textit{corpus error level}.

\subsection{Noising the Data}
When noising an input sentence, we first sample a token edit probability from the error amount distribution, scaled according to the required number of token edits.
We then introduce the desired aspects with the chosen error level.

\looseness1
We allowed the framework to generate any noising aspect, including adding new tokens, in test sets without token-level gold annotations: neural machine translation, GLUE benchmark, tokens outside named entities in NER and tokens outside the answer in reading comprehension.

\looseness1
When introducing errors into classification test sets with token-level gold annotations, we need to maintain the original tokenization. For this reason, we allowed only the first 4 aspects for the following data: morpho-syntactic analysis, tokens inside named entity spans in NER and tokens inside answers in the reading comprehension task.

All experiments are repeated with 5 different random seeds and we report means with standard deviations.

\section{Evaluated Tasks}


\subsection{Morpho-syntactic Analysis}

\noindent\textbf{Model} We employed UDPipe \cite{straka2019evaluating}, a~tool for morpho-syntactic analysis.

\noindent\textbf{Dataset} We used the Universal Dependencies 2.3 \cite{ud23} corpus (UD
2.3).\footnote{Many English UD test set tokens contain casing or spelling errors, propagated into lemmas, rendering such data unsuitable for analysis. We try to use only error-free English test documents and we therefore drop all test documents containing a sentence starting with a lowercase character, keeping more than half of the data. Apart from the lemmatization accuracy, the results for full test set are nearly identical.}

\noindent\textbf{Metrics} We utilized the following metrics \cite{zeman-etal-2018-conll} -- \textbf{UPOS}: coarse POS tags accuracy, \textbf{UFeats}: fine-grained morphological features accuracy, \textbf{Lemmas}: lemmatization accuracy, \textbf{LAS}: labeled attachment score and \textbf{MLAS}: combination of morphological tags and syntactic relations.

\subsection{Named Entity Recognition}

\noindent\textbf{Model} Recently published architecture \cite{strakova-etal-2019-neural} was used for NER evaluation.

\noindent\textbf{Dataset} For English and German, we evaluated on the standard CoNLL-2003 shared task data \cite{tjong-kim-sang-de-meulder-2003-introduction}; for Czech, we used a fine-grained Czech Named Entity Corpus 2.0 \cite{sevcikova2007cnec} with 46 types of nested entities.

\noindent\textbf{Metric} The evaluation metric is F1 score.

\subsection{Neural Machine Translation}
\label{ssec:nmt_fourth}

\noindent\textbf{Model}
We chose a state-of-the-art Czech-to-English NMT system CUBBITT \citep{popel2020transforming},
 but we trained it on the newest version (2.0) of the CzEng parallel corpus \citep{kocmi2020announcing}.
We trained with batch size of ca. 23k tokens for 550k steps, saved a checkpoint each hour (ca. 4600 steps) and selected the checkpoint with the highest dev-set BLEU (which was at 547k steps).

\noindent\textbf{Dataset}
We use WMT17 (newstest2017, 3005 sentences)\footnote{\url{http://statmt.org/wmt17}} as our development set.
Our test set is a concatenation of WMT13, WMT16 and WMT18 (8982 sentences in total).

\noindent\textbf{Metric}
We evaluate the translation quality with case-insensitive BLEU score.\footnote{
 We use SacreBLEU \citep{post-2018-call} with signature BLEU\kern-.09em+\kern-.09em case.lc\kern-.09em+\kern-.09em numrefs.1\kern-.09em+\kern-.09em smooth.exp\kern-.09em+\kern-.09em tok.intl\kern-.09em+\kern-.09em version.1.4.14.
  When using the case-sensitive version, the results show similar trends,
   except for the Casing aspect, which causes more harm to the score,
   as could be expected.
  However, it is questionable if copying the ``wrong'' casing to the translation (e.g. not capitalizing the first word in a sentence
  or using all-uppercase) should be considered a translation error.
  We thus opted for case-insensitive BLEU as our primary metric.
}

\subsection{GLUE Benchmark}
\label{sec:glue}

We select a subset of GLUE~\cite{wang2018glue} tasks, namely Microsoft Research Paraphrase Corpus (MRPC), Semantic Textual Similarity Benchmark (STS-B), Quora Question Pairs (QQP) and The Stanford Sentiment Treebank (SST-2). We finetune BERT on each of these tasks and evaluate them on various levels of noise.

\noindent\textbf{Model} We finetune pretrained BERT with an additional feed-forward neural network with one hidden layer predicting score on particular task's data. We use \textit{bert-base-cased} configuration and HuggingFace's Transformers~\cite{Wolf2019HuggingFacesTS} implementation.

\noindent\textbf{Dataset} We use official GLUE datasets as provided by \url{https://gluebenchmark.com/tasks}.

\noindent\textbf{Metric} We report following metrics: F1 for MRPC and QQP, Pearson-Spearman Corr for STS-B and accuracy for SST-2.

\subsection{Reading Comprehension}
\label{sec:qa}

\noindent\textbf{Model} We utilize a BERT base architecture with a standard SQuAD classifier on top~\citep{devlin2018bert}.

\noindent\textbf{Dataset} We employ English SQuAD 2~\citep{SQuAD2} and its
Czech translation~\citep{mackova-straka-2020}.

\noindent\textbf{Metric} Our experiments are evaluated using F1 score.





\section{Robustness to Noise}

\begin{figure*}[t!]
    \newcommand{\subcapsize}{\fontsize{8.5pt}{10pt}\selectfont}
    \centering
    \begin{subfigure}{.49\hsize}
        \centering
        \includegraphics[width=\hsize]{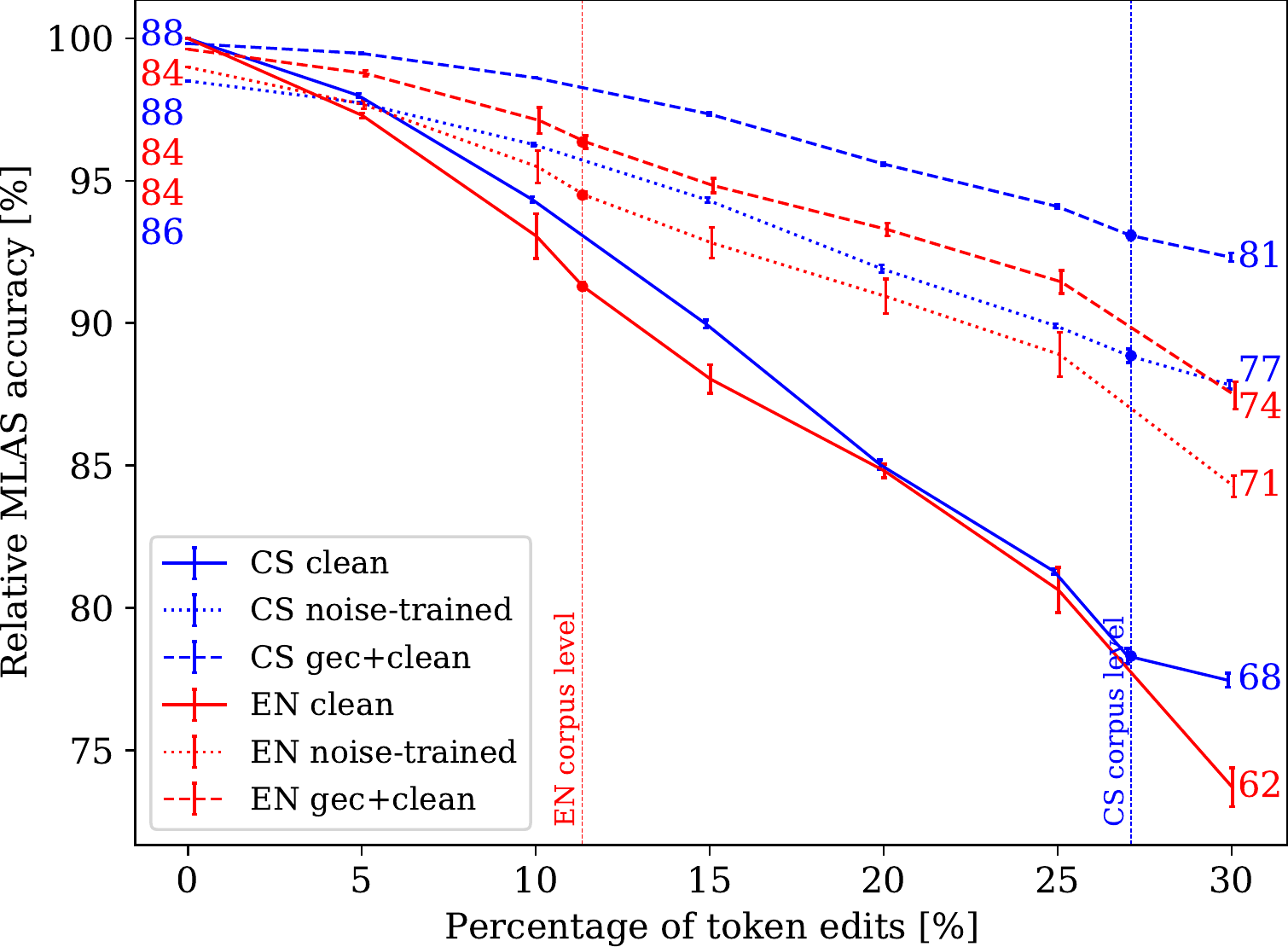}
        \caption{\subcapsize Morpho-syntactic analysis (relative MLAS), Second Learners\vspace{2em}}
        \label{fig:udpipe_corr_mlas}
    \end{subfigure}
    \begin{subfigure}{.49\hsize}
        \centering
        \includegraphics[width=\hsize]{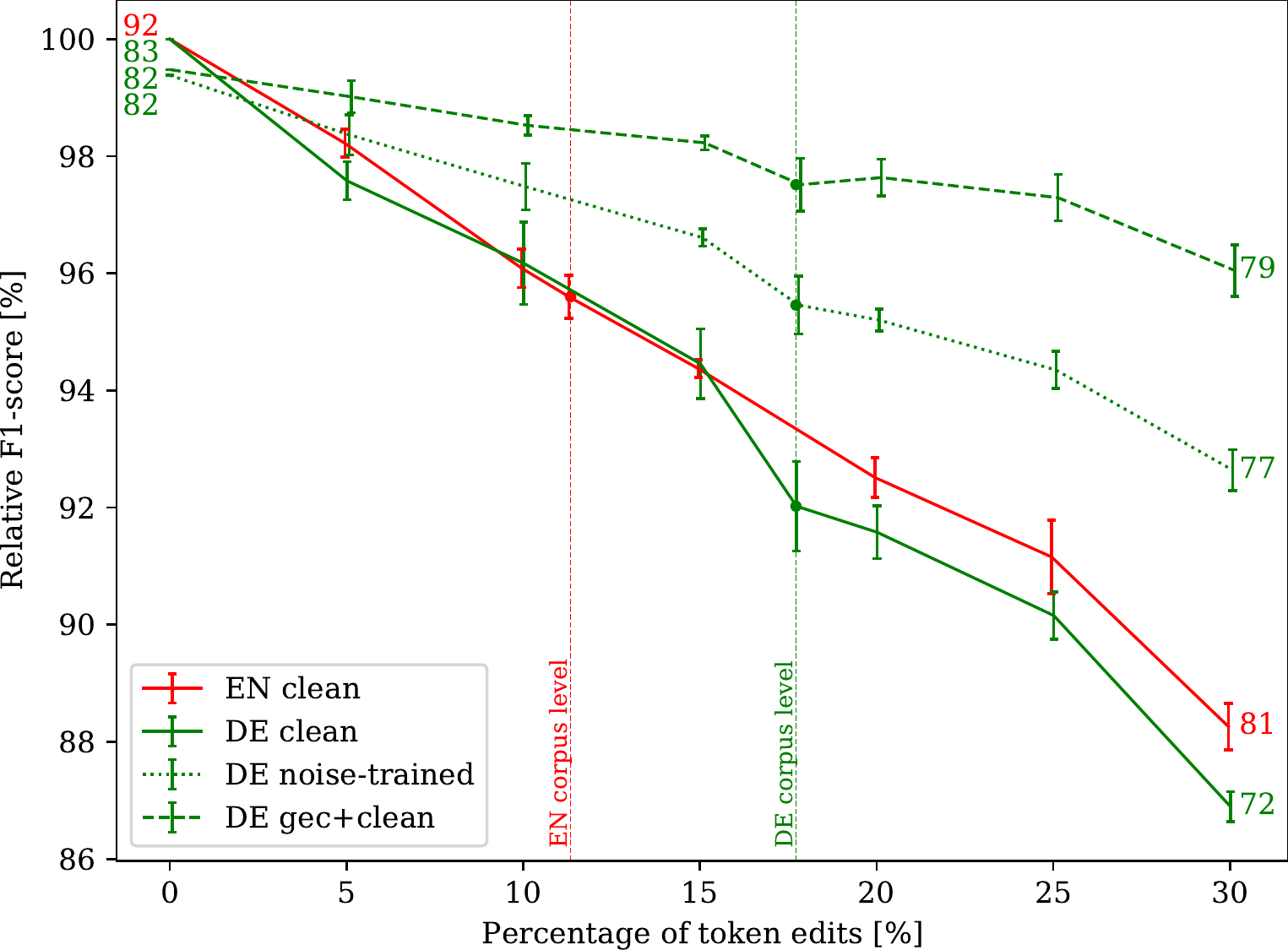}
        \caption{\subcapsize NER (relative F1), Second Learners\vspace{2em}}
        \label{figure_ner_xfreq_cor}
    \end{subfigure}
    \begin{subfigure}{.49\hsize}
        \centering
        \includegraphics[width=\hsize]{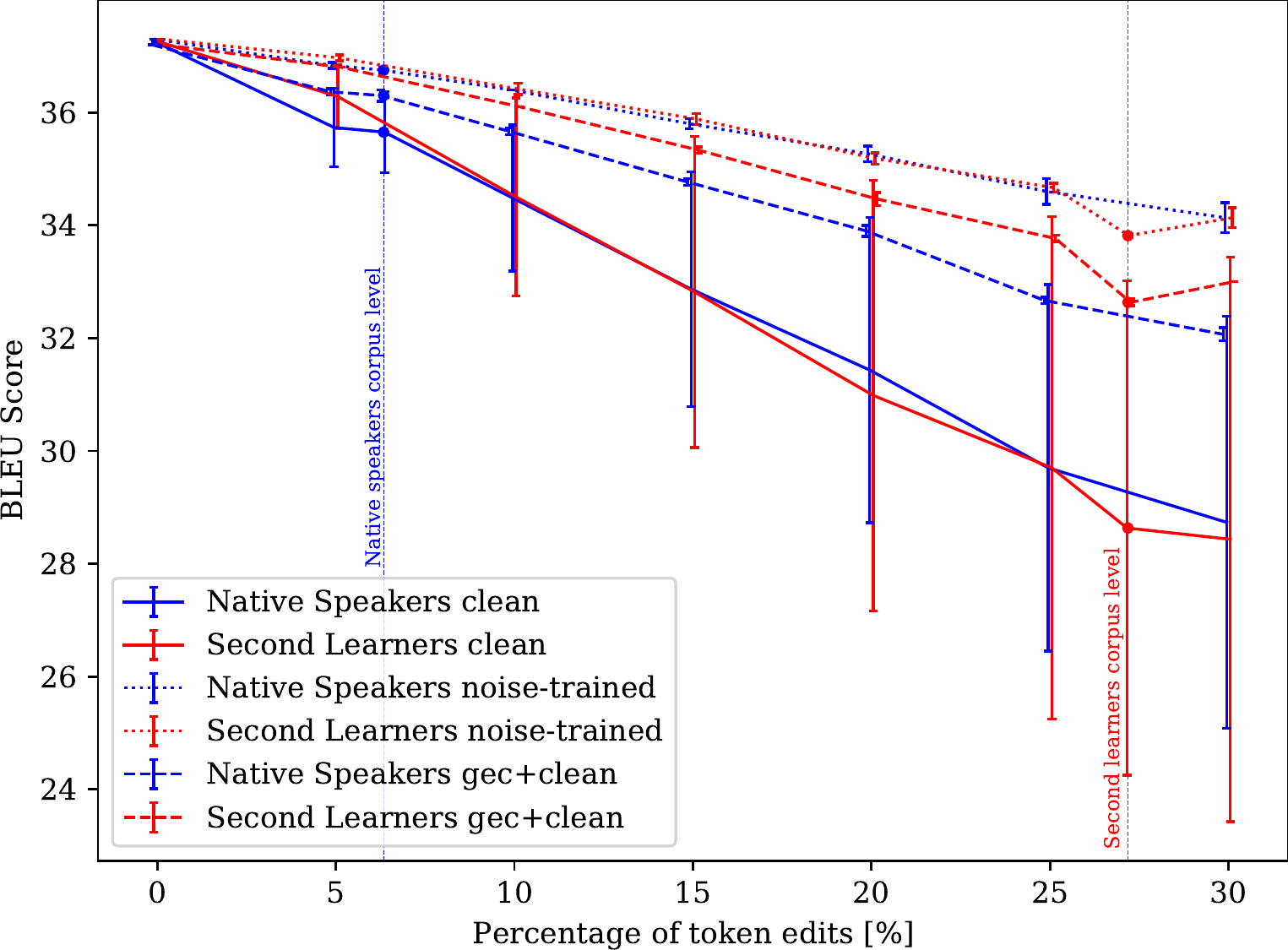}
        \caption{\subcapsize NMT (BLEU), Czech Native Speakers and Second Learners\vspace{.5em}}
    \label{figure_nmt_cor}
    \end{subfigure}
    \begin{subfigure}{.49\hsize}
        \centering
        \includegraphics[width=.99\hsize]{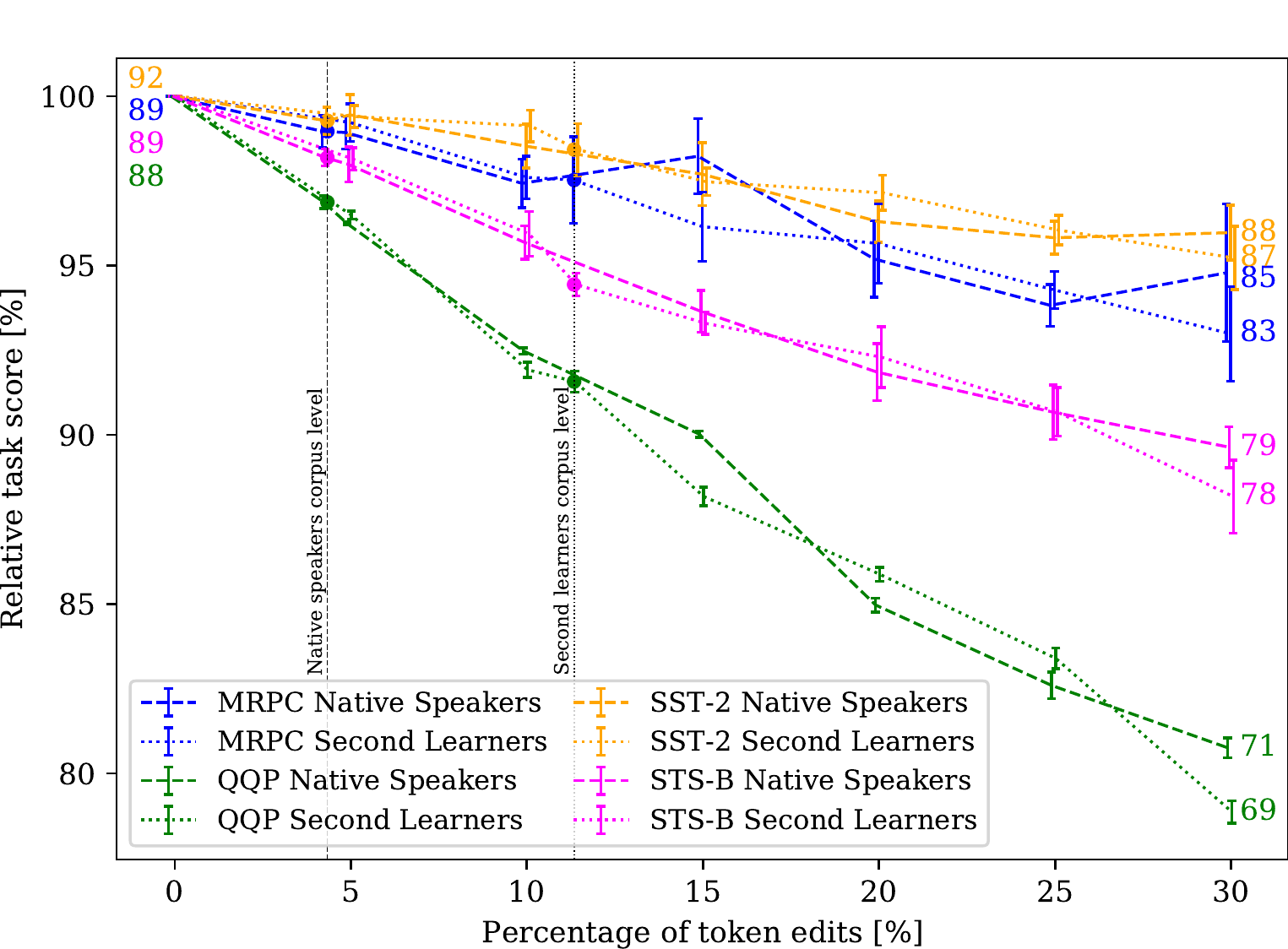}
        \caption{GLUE (relative task score), English Native Speakers and Second Learners\vspace{.5em}}
        \label{fig:glue}
    \end{subfigure}
    \caption{Increasing percentage of token edits with clean model, noise-trained model and grammatical-error-correction. Numbers near lines are absolute values.\vspace{.8em}}
    \label{fig:token_edits}
\end{figure*}

\begin{figure*}[t!]
    \centering
    \vspace{2em}
    \includegraphics[width=\hsize]{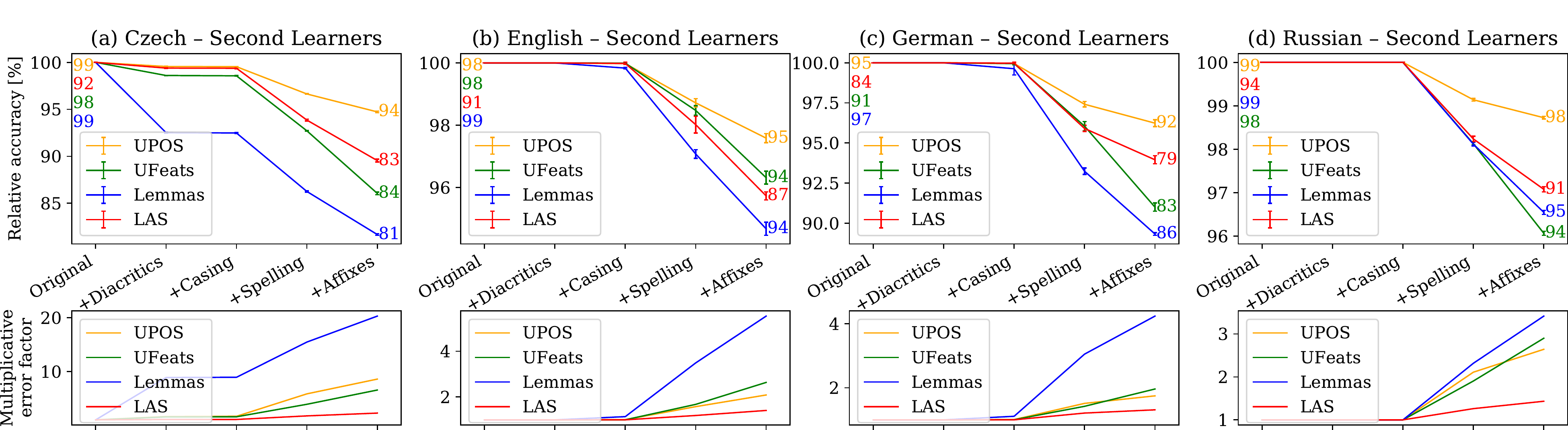}    
    \vspace{.2em}
    \caption{Morpho-syntactic tasks with additive noising aspects in Second Learners profile across (a)~Czech, (b)~English, (c)~German and (d)~Russian. The amount of introduced errors is the corpus error level for each aspect. \textbf{Upper~row}~Accuracy relative to original data accuracy, numbers near lines are absolute values. \textbf{Lower row}~Multiplicative factor of errors to original data error rate.}
    \label{figure_udpipe_apects}
\end{figure*}

\begin{figure}[t!]
    \centering
    \begin{subfigure}{\hsize}
        \centering
        \includegraphics[width=\hsize]{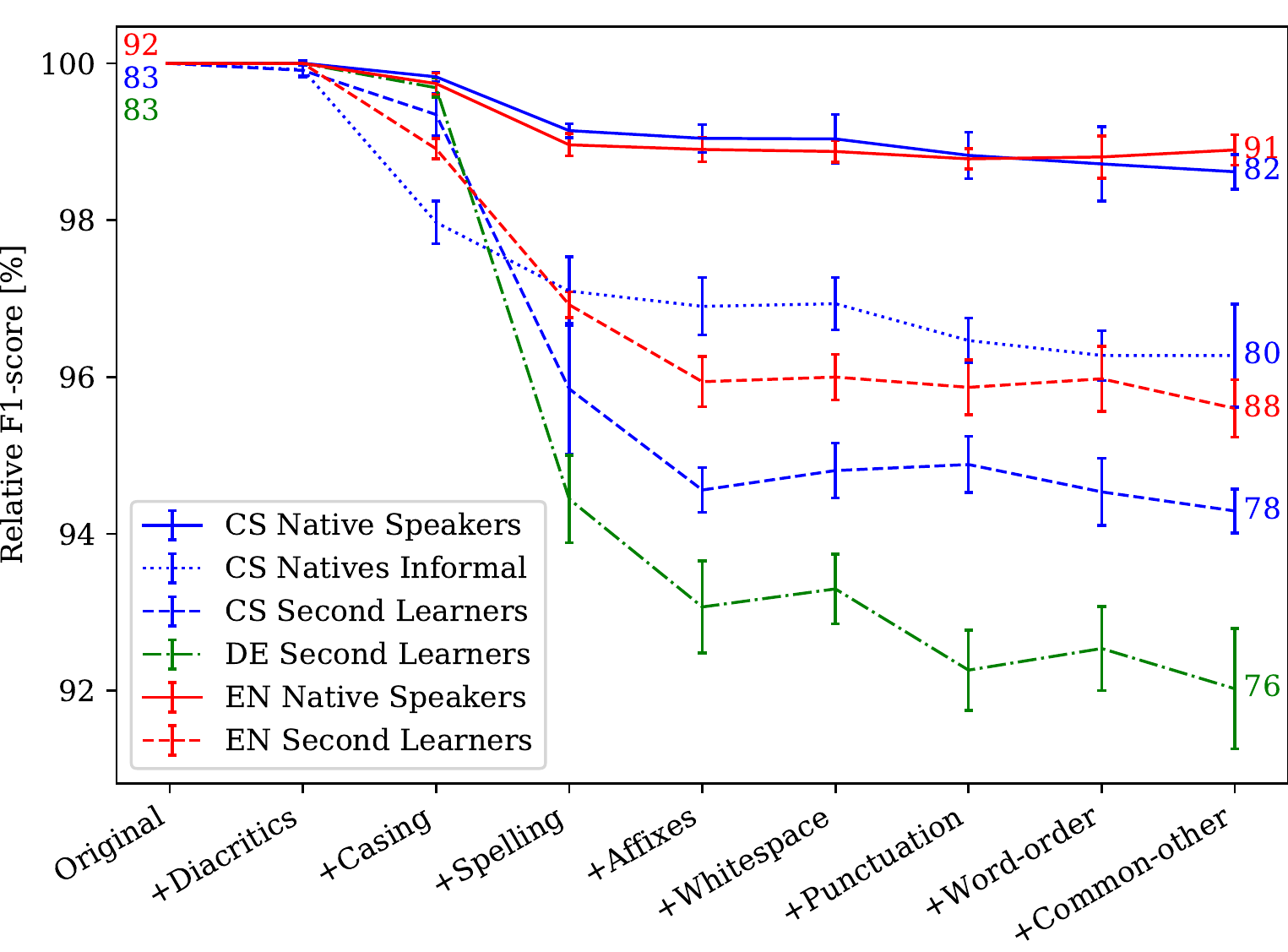}
        \caption{NER (relative F1), selected profiles in CS, EN and DE\vspace{1.8em}}
        \label{figure_ner_xaspects}
    \end{subfigure}
    \begin{subfigure}{\hsize}
        \centering
        \includegraphics[width=\hsize]{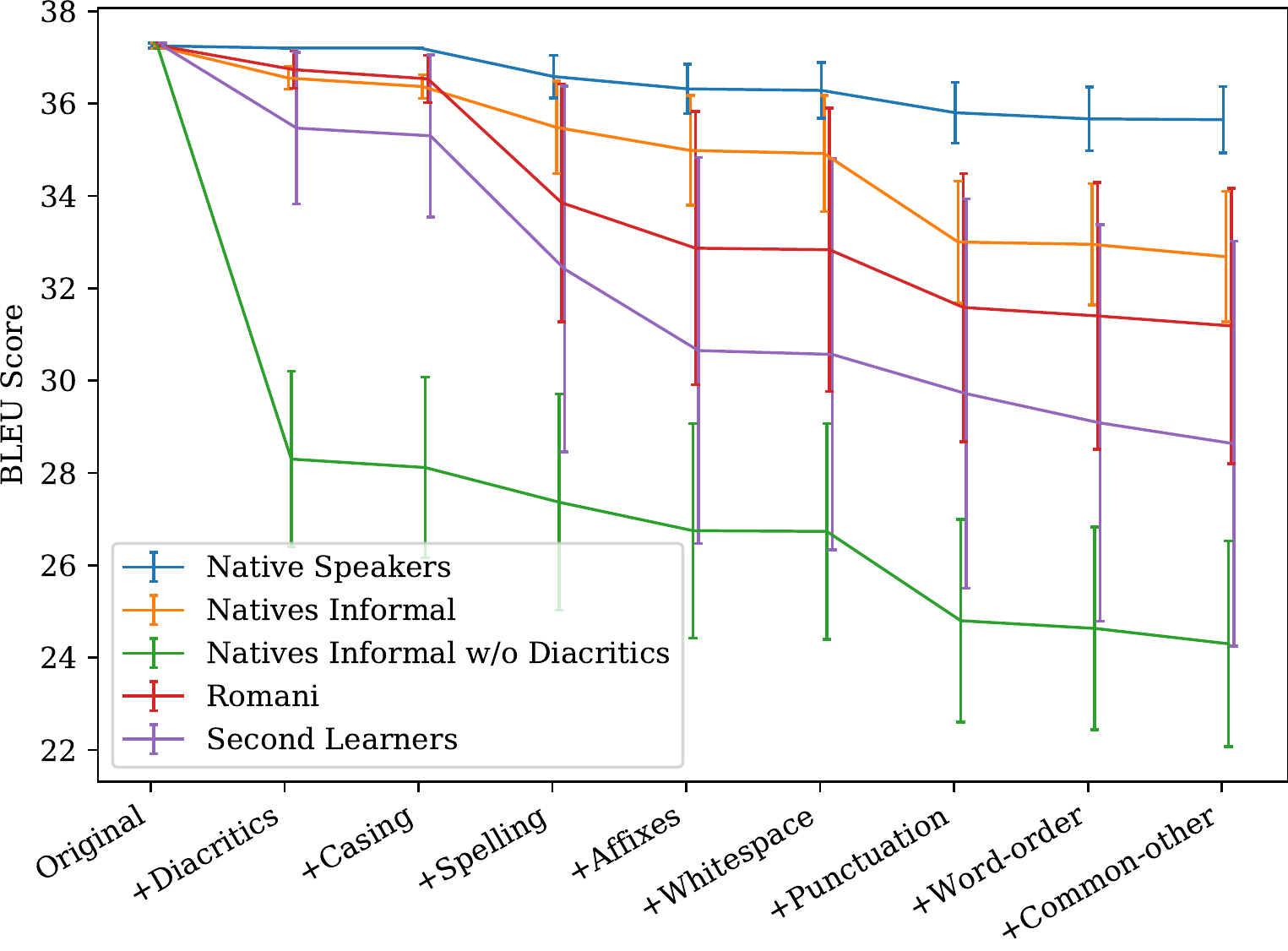}
        \caption{NMT (BLEU), all Czech profiles\vspace{1.8em}}
        \label{figure_nmt_fourth}
    \end{subfigure}
    \begin{subfigure}{\hsize}
    \centering
    \includegraphics[width=.99\hsize]{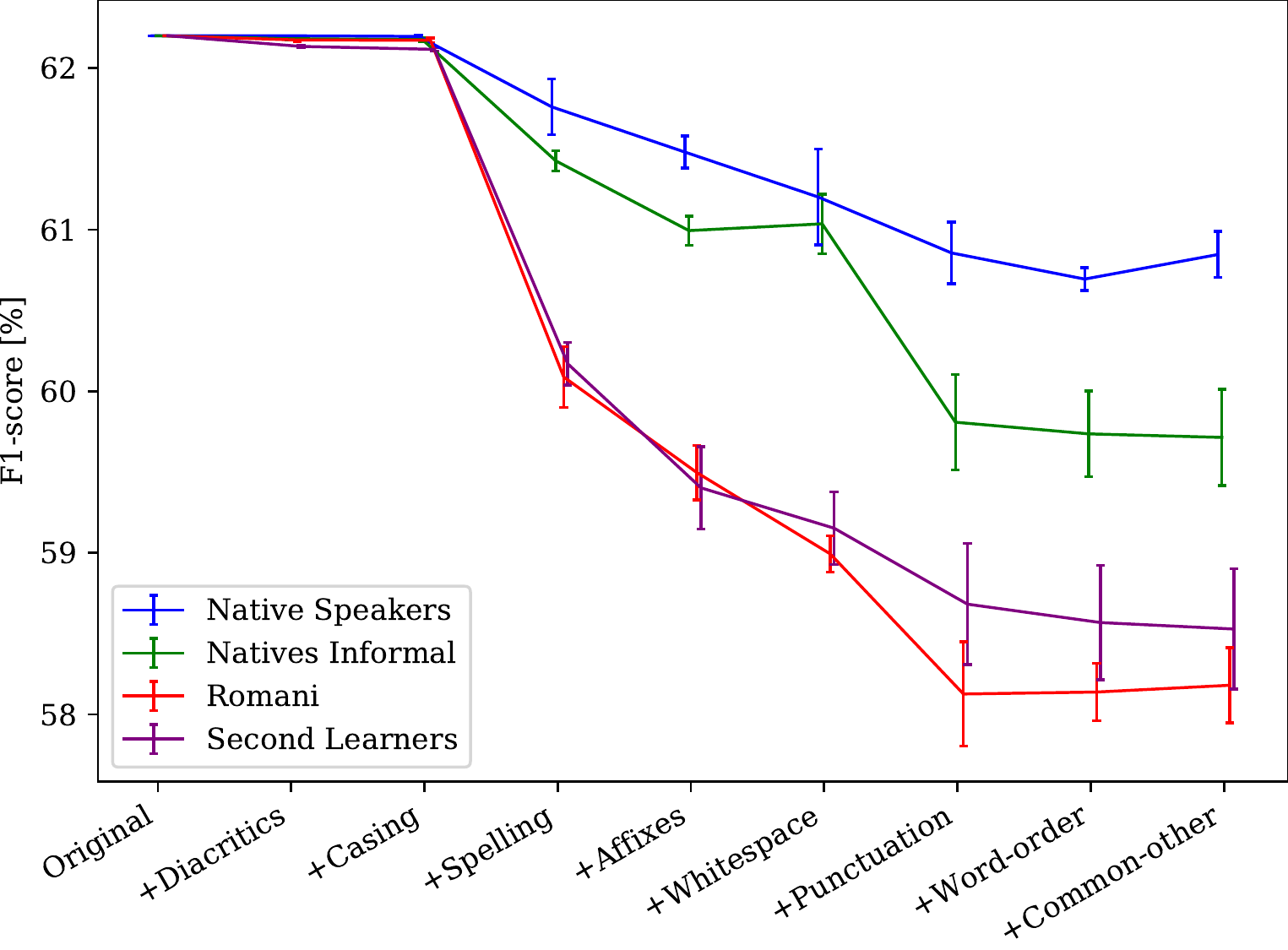}
    \caption{Reading comprehension (relative F1), all Czech profiles\vspace{.8em}}
    \label{fig:squad}
    \end{subfigure}
    
    \caption{Evaluation with additive noising aspects. The amount of introduced errors is the corpus error level for each aspect. Numbers near lines are absolute values. }
    \label{fig:aspects}
\end{figure}

We evaluated the models robustness both to the amount of noise (Figure~\ref{fig:token_edits}) and to error types (Figures~\ref{figure_udpipe_apects} and \ref{fig:aspects}).

\begin{figure}
    \centering
    \includegraphics[width=\hsize]{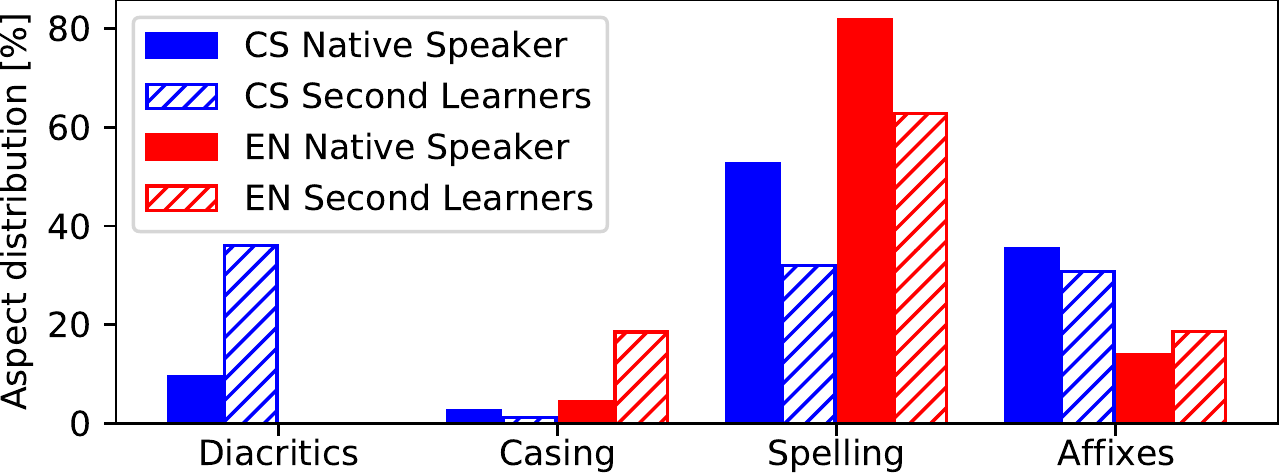}
    \caption{Proportional distribution of the first 4 aspects (diacritics, casing, spelling, affixes) in Czech and English.}
    \label{fig:aspect_ratios}
\end{figure}

A unifying trend can be observed in models performance with respect to increasing percentage of token edits. Solid lines in Figure~\ref{fig:token_edits} display the morpho-syntactic MLAS, NER F1 and NMT BLEU on texts with up to $30\%$ of token edits. The relative performance decreases roughly linearly with the amount of token edits, in accordance with previous findings \cite{rychalska2019models}. The tendency is consistent across tasks, languages and profiles: For example, compare the Czech and English Second Learners profiles in morpho-syntactic analysis (Figure~\ref{fig:udpipe_corr_mlas}) or Czech Native Speakers and Czech Second Learners profiles in the NMT clean model (Figure~\ref{figure_nmt_cor}), which exhibit similar behaviour despite their differing distributions of aspects (Figure~\ref{fig:aspect_ratios}). This consistency implies that it is the sheer amount of noise rather than the distribution of aspects, that contributes to the model performance deterioration. More results are available in Supplementary Material (Figures~\ref{supp:fig:udpipe_corr_lemmas} and \ref{supp:fig:ner_csen}).

Estimating the amount of noise is important, as the \textit{corpus error level} differs greatly across languages and profiles. For example, compare the Second Learners profile in English ($11.3\%$ token edits) and Czech ($27.1\%$) in Figure~\ref{fig:udpipe_corr_mlas}, or in Czech, see Native Speakers ($6.4\%$) and Second Learners ($27.1\%$) in Figure~\ref{figure_nmt_cor}. Testing near the estimated noisiness level provides more accurate evaluation of the models' performance.

From a qualitative point of view, spelling and affixes make for the major performance drop in morpho-syntactic analysis (Figure~\ref{figure_udpipe_apects}), NER (Figure~\ref{figure_ner_xaspects}) and NMT (Figure~\ref{figure_nmt_fourth}). 

Some tasks are more sensitive to certain aspects: Casing is a crucial aspect for NER. This is clearly shown in the Czech Natives Informal profile, which contains text scraped from the internet discussions and contains nontrivial amount of casing errors (Figure~\ref{figure_ner_xaspects}). We further elaborate the casing aspect effect on NER in Section~\ref{supp:sec:ner} in Supplementary Material. In NMT and reading comprehension, errors in punctuation seem to decrease the model performance consistently across all profiles (Figures~\ref{figure_nmt_fourth} and \ref{fig:squad}, respectively).

For Czech as a language with diacritic marks, diacritics is an interesting aspect. We can see that when it is introduced at a \textit{corpus error level}, the Czech model's performance on Lemmas drops by circa 7 percent. Figure~\ref{supp:fig:udpipe_diacritics} in Supplementary Material further illustrates that performance significantly deteriorates when all diacritics is stripped, which is quite common in informal Web texts. Similarly, to emphasize the effect of the diacritization aspect on NMT, we created a new profile \textit{Natives Informal w/o Diacritics} from the Natives Informal profile by stripping all diacritization. Figure~\ref{figure_nmt_fourth} shows that not using diacritics at all results in a performance drop of ca. 10 BLEU points.

Some tasks are more sensitive to noise than others. Lemmatization is the most sensitive to errors (20 times more errors when processing Czech Second Learners texts with a clean model, see Figure~\ref{figure_udpipe_apects}), which is understandable, given that all lemma characters must be generated correctly from a corrupted surface token. The effect on POS tagging is the least pronounced (Figure~\ref{figure_udpipe_apects}), although 8 times as many errors in Czech (when processing noisy texts with a clean model) makes the POS tags much less reliable.

\section{Noise-coping Strategies}
\label{sec:noise-coping}

We implemented and evaluated two strategies to alleviate the performance drop on noisy inputs: \textit{external} and \textit{internal} correction. In the \textit{external} correction approach, we use a separately trained grammatical-error-correction model to denoise texts before inputting them to the model itself. In the \textit{internal} correction approach, we instead directly train the model on a combination of noisy and authentic texts. 

We hypothesise that the external approach may be better in scenarios with small amount of annotated data. In such cases, only few iterations over training data are typically performed to prevent overfitting, and we suppose that learning the task itself and denoising at the same time would harm its performance a lot. Contrarily, with enough data and appropriate model capacity, learning the denoising and the task jointly may reduce the amount of potential false positives that might be otherwise proposed by the external language corrector.

\subsection{External Correction Model}

We use the grammatical-error-correction system of \newcite{naplava2019grammatical} in our experiments. Their models trained on Czech, German and Russian achieve state-of-the-art results and slightly below state-of-the-art results on English. We use their ``pretrained'' version.

We modified the pipeline of \citet{naplava2019grammatical} to train on detokenized text. Furthermore, we also trained new grammatical-error-correction models which only make corrections that strictly keep the given tokenization (important in morpho-syntactic annotations). To sum up, we trained two types of grammatical-error-correction models: 1. detokenized error correction model (for NMT) 2. tokenization-preserving error-correction model (for morpho-syntactic tasks and NER).

\subsection{Training on Noisy Data}

In the \textit{internal} approach to increase model robustness, we train the systems on a mixture of original and noisy data, while keeping the number of training steps unchanged. The noisy data are generated using the \MLNoise{} framework operating on development profiles and concatenated to original data.

\begin{figure}
    \centering
    \includegraphics[width=\hsize]{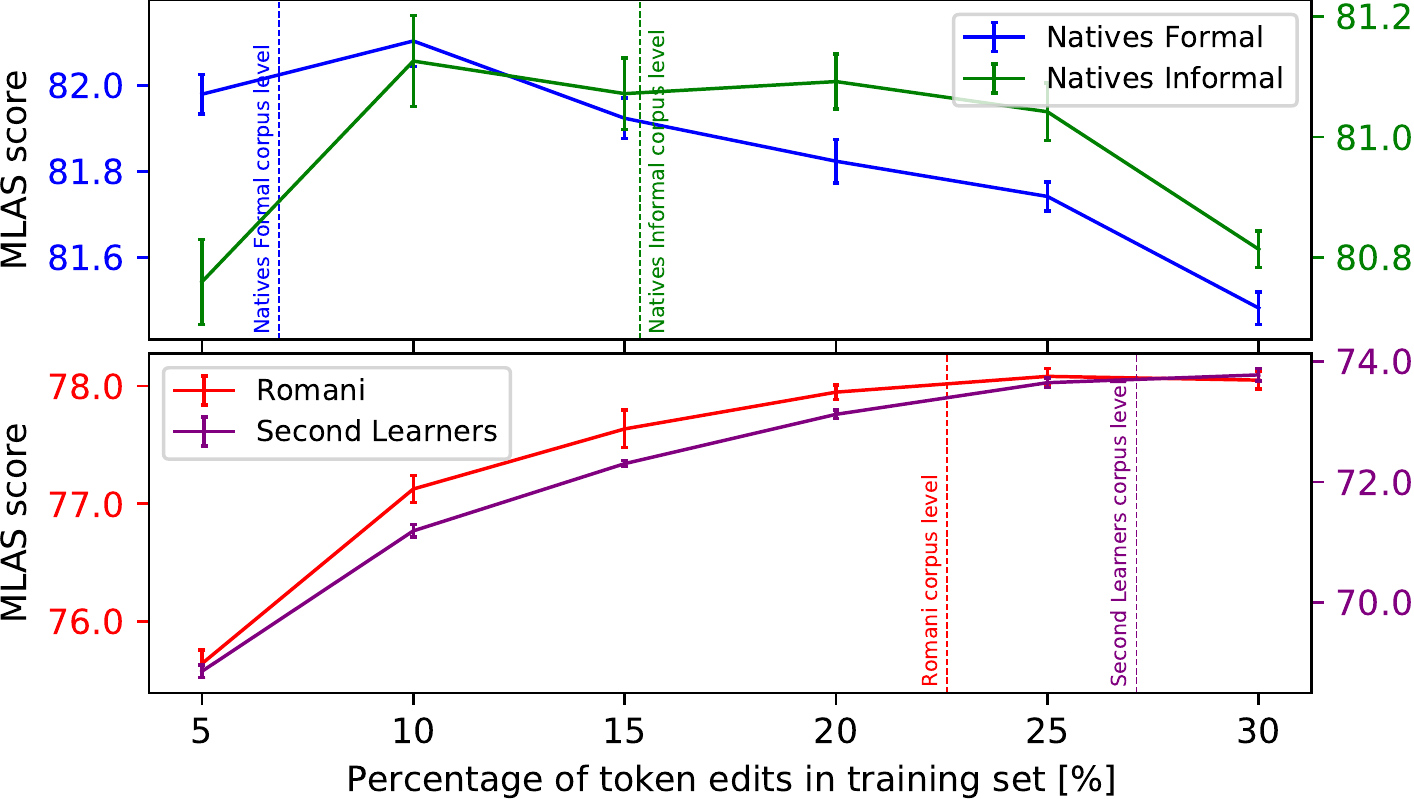}
    \caption{Morpho-syntactic analysis: Training data increasingly noised with each single profile, evaluation with the corresponding profile corpus error level.}
    \label{figure_udpipe_training_amount}
\end{figure}
\looseness-1
We noise the training data with appropriately estimated corpus error levels in all our experiments. 
To illustrate the effect of noise level introduced into training data, we trained the UDPipe on variably noised morpho-syntactic data for all four Czech profiles. In each single profile, we increasingly noised the morpho-syntactic training data and evaluated on the testing data noised with the corresponding profile \textit{corpus error level}. In all cases, the best performance is found near the corpus error level (Figure~\ref{figure_udpipe_training_amount}).

When training the NMT model, the best checkpoint on a development set consisting of concatenated standard WMT17 and WMT17 noised with our framework is selected. 

We train a single model for each language on a concatenation of noisy data generated by all profiles of the particular language. This makes the final model generalize well across all profiles, although training a single model for each profile could make sense for other scenarios.

\subsection{Evaluation}
 
\begin{figure}
    \centering
    \includegraphics[width=\hsize]{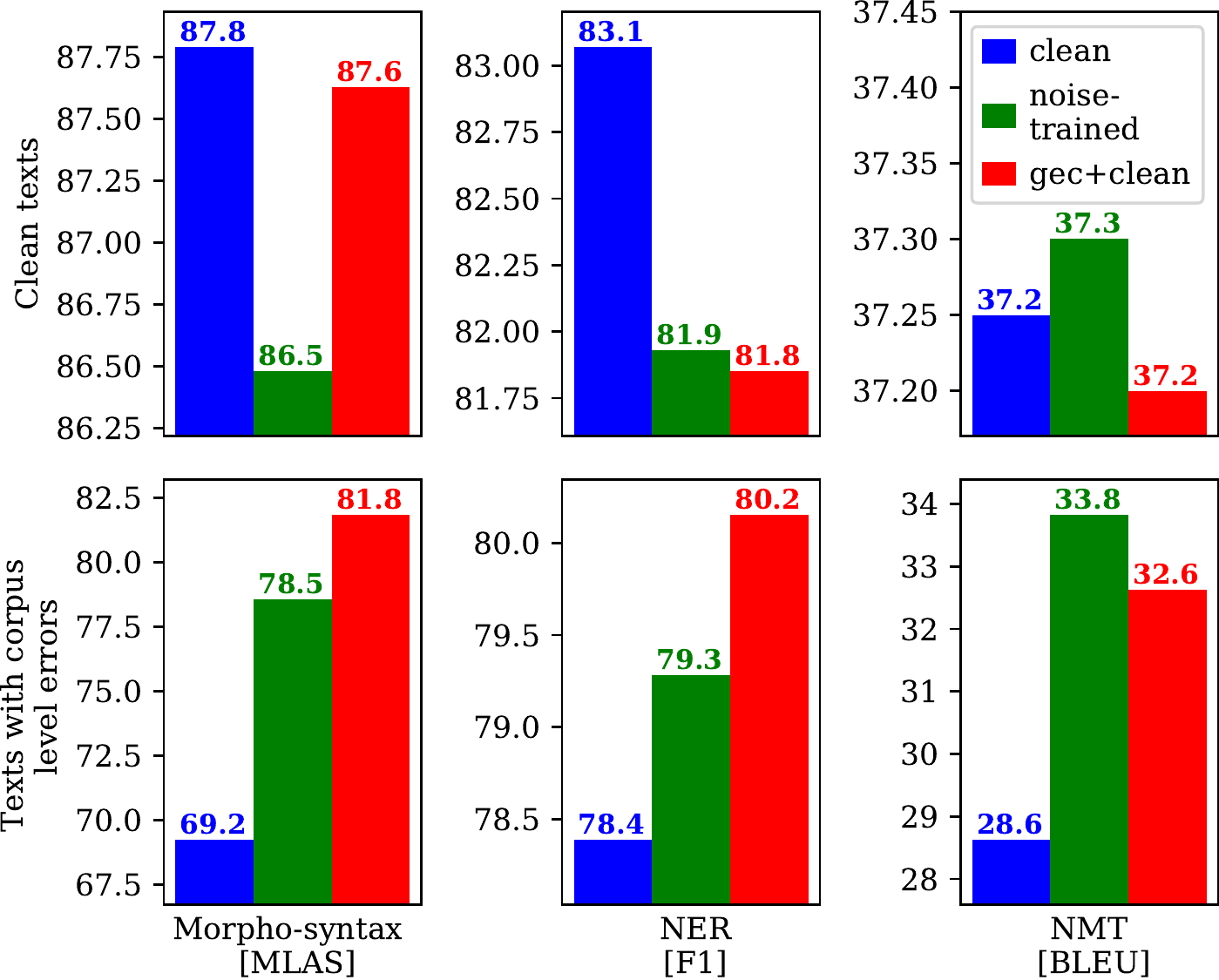}
    \caption{Comparison of three models (clean, noise-trained and GEC-preprocessed) on three tasks in Czech Second Learners profile. \textbf{Upper~row}~Original clean test data. \textbf{Lower row}~Test data with corpus error level noise.}
    \label{figure_cor_comparison}
\end{figure}

We present the effect of both the \textit{internal} and \textit{external} noise-coping strategies in Figure~\ref{fig:token_edits}. 
There are two main points of interest in the graphs: the first one showing performance of models on clean texts and the second one showing model performance on texts with corpus level errors. Additionally, an excerpt showing performance of Czech Second Learners on these two levels is presented in Figure~\ref{figure_cor_comparison}. 

It is not a surprise that the model trained on clean training data surpasses the noise-coping models on the clean test data. Adapting to noise clearly comes with a cost. Surprisingly though, the clean model head start is only marginal in the NMT task.

The clean models perform substantially worse than either of the two proposed methods in all three tasks when errors are introduced in the same amount as the corpus error level (marked with vertical lines in Figure~\ref{fig:token_edits}). Therefore, whenever noisy inputs of particular domain are expected, it is beneficial to adapt to noise using either of the two methods.

With increasing noise, the gap between the clean model and the \textit{external} and \textit{internal} model grows in all three tasks (Figure~\ref{fig:token_edits}). There is a threshold at which the noise-coping models surpass the clean model for each task. Interestingly, the threshold oscillates around relatively low noise levels up to $5\%$ of token edits.

Finally, we confirm our initial hypothesis that \textit{external} approach with GEC model works better than \textit{internal} approach on low resource tasks: morpho-syntactic analysis and named-entity recognition. The \textit{internal} approach then outperforms \textit{external} approach on machine translation task for which there is a large amount of training data and a model with greater capacity.

\begin{table}
 \centering
 \setlength{\tabcolsep}{12pt}
 \begin{tabular}{l c c}\toprule
 \multirow{2}{*}{System} & \multicolumn{2}{c}{BLEU on Faust}\\ \cmidrule{2-3}
                   & Noisy & Cleaned \\\midrule 
 clean             & 43.3  & 50.9 \\ 
 noise-trained     & 47.0  & 50.5 \\ 
 gec+clean         & 44.1  & 50.4 \\ 
 \bottomrule
    \end{tabular}
    \caption{NMT results on 
    authentic user noisy texts.
    We report BLEU on the Faust-Noisy test set with noisy input sentences and also on Faust-Cleaned that has manually corrected sentences on input.}
    \label{table:faust}
\end{table}

\section{Evaluating on Authentic User Text}
\label{sec:authentic}

We assembled a new dataset for MT evaluation consisting of 2223 authentic Czech noisy input sentences translated into English, which we release at \url{http://hdl.handle.net/11234/1-3775}.
The sentences originate from the project FAUST\footnote{\url{https://ufal.mff.cuni.cz/grants/faust}}
 where they were collected from various users of \url{reverso.net}.
The advantage of this dataset is that in addition to the original Czech noisy sentences,
 there are manually corrected Czech sentences
 and manual translations to English.


On this dataset, we evaluate our neural machine translation models from Section~\ref{ssec:nmt_fourth} and Section~\ref{sec:noise-coping}, specifically
 the \textit{clean} model trained on clean data,
 \textit{noise-trained} model trained on a mixture
 of authentic and noised data and their combination with external grammatical-error-correction system.
The results of these systems on authentic noisy texts are presented in Table~\ref{table:faust}. It is evident that noise-trained model outperforms clean model by a large margin on Faust-Noisy data while not losing much precision on Faust-Cleaned data. Similarly to our conclusions in Section~\ref{sec:noise-coping}, the external grammatical-error-correction system helps the clean model on noisy data, however is inferior to noise-trained model.

\section{Conclusions}

We estimated natural error probabilities statistically from real-world grammatical-error-correction corpora in order to model and generate noisy inputs for machine learning tasks. We extensively evaluated several state-of-the-art NLP downstream systems with respect to their robustness to input noise, both in increasing level of text noisiness and in variations of error types. We confirmed that the noise hurts the model performance substantially and we compared two coping strategies: training with noise and preprocessing with GEC, concluding that each strategy is beneficial in different scenarios. Finally, we also presented authentic noisy data evaluation using a newly assembled dataset for machine translation with authentic Czech noisy sentences translated to English. We release both the new framework (under MPL 2.0) at \url{https://github.com/ufal/kazitext} and the newly assembled dataset (under CC BY-NC-SA license) at \url{http://hdl.handle.net/11234/1-3775}.


\section*{Acknowledgements}

This work has been supported by the Grant Agency of the Czech Republic, project EXPRO LUSyD (GX20-16819X).
This research was also partially supported by SVV project number 260 575 and GAUK 578218 of the Charles University.
It has has also been using data provided by the LINDAT/CLARIAH-CZ Research Infrastructure (https://lindat.cz), supported by the Ministry of Education, Youth and Sports of the Czech Republic (Project No. LM2018101).


\bibliographystyle{acl_natbib}
\bibliography{w-nut_kazitext_2021}

\begin{thebibliography}{38}
\expandafter\ifx\csname natexlab\endcsname\relax\def\natexlab#1{#1}\fi

\bibitem[{Belinkov and Bisk(2017)}]{belinkov2017synthetic}
Yonatan Belinkov and Yonatan Bisk. 2017.
\newblock Synthetic and natural noise both break neural machine translation.
\newblock \emph{arXiv preprint arXiv:1711.02173}.

\bibitem[{Boyd(2018)}]{boyd2018using}
Adriane Boyd. 2018.
\newblock Using wikipedia edits in low resource grammatical error correction.
\newblock In \emph{Proceedings of the 2018 EMNLP Workshop W-NUT: The 4th
  Workshop on Noisy User-generated Text}, pages 79--84.

\bibitem[{Choe et~al.(2019)Choe, Ham, Park, and Yoon}]{choe2019neural}
Yo~Joong Choe, Jiyeon Ham, Kyubyong Park, and Yeoil Yoon. 2019.
\newblock A neural grammatical error correction system built on better
  pre-training and sequential transfer learning.
\newblock In \emph{Proceedings of the Fourteenth Workshop on Innovative Use of
  NLP for Building Educational Applications}, pages 213--227.

\bibitem[{Dahlmeier et~al.(2013)Dahlmeier, Ng, and Wu}]{dahlmeier2013building}
Daniel Dahlmeier, Hwee~Tou Ng, and Siew~Mei Wu. 2013.
\newblock \href {https://www.aclweb.org/anthology/W13-1703} {Building a large
  annotated corpus of learner english: The nus corpus of learner english}.
\newblock In \emph{Proceedings of the eighth workshop on innovative use of NLP
  for building educational applications}, pages 22--31.

\bibitem[{Devlin et~al.(2019)Devlin, Chang, Lee, and
  Toutanova}]{devlin2018bert}
Jacob Devlin, Ming-Wei Chang, Kenton Lee, and Kristina Toutanova. 2019.
\newblock Bert: Pre-training of deep bidirectional transformers for language
  understanding.
\newblock In \emph{Proceedings of the 2019 Conference of the North American
  Chapter of the Association for Computational Linguistics: Human Language
  Technologies, Volume 1 (Long and Short Papers)}, pages 4171--4186.

\bibitem[{Ge et~al.(2018)Ge, Wei, and Zhou}]{ge2018reaching}
Tao Ge, Furu Wei, and Ming Zhou. 2018.
\newblock Reaching human-level performance in automatic grammatical error
  correction: An empirical study.
\newblock \emph{arXiv preprint arXiv:1807.01270}.

\bibitem[{Glockner et~al.(2018)Glockner, Shwartz, and
  Goldberg}]{glockner2018breaking}
Max Glockner, Vered Shwartz, and Yoav Goldberg. 2018.
\newblock Breaking nli systems with sentences that require simple lexical
  inferences.
\newblock In \emph{Proceedings of the 56th Annual Meeting of the Association
  for Computational Linguistics (Volume 2: Short Papers)}, pages 650--655.

\bibitem[{Granger(1998)}]{granger1998}
Sylviane Granger. 1998.
\newblock {The computer learner corpus: A versatile new source of data for SLA
  research.}
\newblock In Sylviane Granger, editor, \emph{{Learner English on Computer}},
  pages 3--18. Addison Wesley Longman, London and New York.

\bibitem[{Heigold et~al.(2018)Heigold, Varanasi, Neumann, and van
  Genabith}]{heigold2018robust}
Georg Heigold, Stalin Varanasi, G{\"u}nter Neumann, and Josef van Genabith.
  2018.
\newblock How robust are character-based word embeddings in tagging and mt
  against wrod scramlbing or randdm nouse?
\newblock In \emph{Proceedings of the 13th Conference of the Association for
  Machine Translation in the Americas (Volume 1: Research Papers)}, pages
  68--80.

\bibitem[{Kasewa et~al.(2018)Kasewa, Stenetorp, and
  Riedel}]{kasewa2018wronging}
Sudhanshu Kasewa, Pontus Stenetorp, and Sebastian Riedel. 2018.
\newblock Wronging a right: Generating better errors to improve grammatical
  error detection.
\newblock In \emph{Proceedings of the 2018 Conference on Empirical Methods in
  Natural Language Processing}, pages 4977--4983.

\bibitem[{Kocmi et~al.(2020)Kocmi, Popel, and Bojar}]{kocmi2020announcing}
Tom Kocmi, Martin Popel, and Ondrej Bojar. 2020.
\newblock Announcing {CzEng} 2.0 parallel corpus with over 2 gigawords.
\newblock \emph{arXiv preprint arXiv:2007.03006}.

\bibitem[{Li et~al.(2019)Li, Michel, Anastasopoulos, Belinkov, Durrani, Firat,
  Koehn, Neubig, Pino, and Sajjad}]{li2019findings}
Xian Li, Paul Michel, Antonios Anastasopoulos, Yonatan Belinkov, Nadir Durrani,
  Orhan Firat, Philipp Koehn, Graham Neubig, Juan Pino, and Hassan Sajjad.
  2019.
\newblock Findings of the first shared task on machine translation robustness.
\newblock In \emph{Proceedings of the Fourth Conference on Machine Translation
  (Volume 2: Shared Task Papers, Day 1)}, pages 91--102.

\bibitem[{Mackov{\'a} and Straka(2020)}]{mackova-straka-2020}
Kate{\v{r}}ina Mackov{\'a} and Milan Straka. 2020.
\newblock Reading comprehension in czech via machine translation and
  cross-lingual transfer.
\newblock In \emph{Text, Speech, and Dialogue}, pages 171--179, Cham. Springer
  International Publishing.

\bibitem[{Michel and Neubig(2018)}]{michel2018mtnt}
Paul Michel and Graham Neubig. 2018.
\newblock Mtnt: A testbed for machine translation of noisy text.
\newblock In \emph{Proceedings of the 2018 Conference on Empirical Methods in
  Natural Language Processing}, pages 543--553.

\bibitem[{N{\'a}plava(2017)}]{naplava2017natural}
Jakub N{\'a}plava. 2017.
\newblock Natural language correction.
\newblock \emph{Diploma Thesis}.

\bibitem[{N{\'a}plava and Straka(2019)}]{naplava2019grammatical}
Jakub N{\'a}plava and Milan Straka. 2019.
\newblock Grammatical error correction in low-resource scenarios.
\newblock In \emph{Proceedings of the 5th Workshop on Noisy User-generated Text
  (W-NUT 2019)}, pages 346--356.

\bibitem[{Nivre et~al.(2018)}]{ud23}
Joakim Nivre et~al. 2018.
\newblock \href {http://hdl.handle.net/11234/1-2895} {Universal dependencies
  2.3}.
\newblock {LINDAT}/{CLARIN} digital library at the Institute of Formal and
  Applied Linguistics ({{\'U}FAL}), Faculty of Mathematics and Physics, Charles
  University.

\bibitem[{Popel et~al.(2020)Popel, Tomkova, Tomek, Kaiser, Uszkoreit, Bojar,
  and {\v{Z}}abokrtsk{\`y}}]{popel2020transforming}
Martin Popel, Marketa Tomkova, Jakub Tomek, {\L}ukasz Kaiser, Jakob Uszkoreit,
  Ond{\v{r}}ej Bojar, and Zden{\v{e}}k {\v{Z}}abokrtsk{\`y}. 2020.
\newblock Transforming machine translation: a deep learning system reaches news
  translation quality comparable to human professionals.
\newblock \emph{Nature Communications}, 11(1):1--15.

\bibitem[{Post(2018)}]{post-2018-call}
Matt Post. 2018.
\newblock \href {https://doi.org/10.18653/v1/W18-6319} {A call for clarity in
  reporting {BLEU} scores}.
\newblock In \emph{Proceedings of the Third Conference on Machine Translation:
  Research Papers}, pages 186--191, Belgium, Brussels. Association for
  Computational Linguistics.

\bibitem[{Rajpurkar et~al.(2018)Rajpurkar, Jia, and Liang}]{SQuAD2}
Pranav Rajpurkar, Robin Jia, and Percy Liang. 2018.
\newblock \href {https://doi.org/10.18653/v1/P18-2124} {Know what you don{'}t
  know: Unanswerable questions for {SQ}u{AD}}.
\newblock In \emph{Proceedings of the 56th Annual Meeting of the Association
  for Computational Linguistics (Volume 2: Short Papers)}, pages 784--789,
  Melbourne, Australia. Association for Computational Linguistics.

\bibitem[{Rei et~al.(2017)Rei, Felice, Yuan, and Briscoe}]{rei2017artificial}
Marek Rei, Mariano Felice, Zheng Yuan, and Ted Briscoe. 2017.
\newblock Artificial error generation with machine translation and syntactic
  patterns.
\newblock In \emph{Proceedings of the 12th Workshop on Innovative Use of NLP
  for Building Educational Applications}, pages 287--292.

\bibitem[{Ribeiro et~al.(2018)Ribeiro, Singh, and
  Guestrin}]{ribeiro2018semantically}
Marco~Tulio Ribeiro, Sameer Singh, and Carlos Guestrin. 2018.
\newblock Semantically equivalent adversarial rules for debugging nlp models.
\newblock In \emph{Proceedings of the 56th Annual Meeting of the Association
  for Computational Linguistics (Volume 1: Long Papers)}, pages 856--865.

\bibitem[{Rozovskaya and Roth(2019)}]{rozovskaya2019grammar}
Alla Rozovskaya and Dan Roth. 2019.
\newblock Grammar error correction in morphologically rich languages: The case
  of russian.
\newblock \emph{Transactions of the Association for Computational Linguistics},
  7:1--17.

\bibitem[{Rozovskaya et~al.(2017)Rozovskaya, Roth, and
  Sammons}]{rozovskaya2017adapting}
Alla Rozovskaya, Dan Roth, and Mark Sammons. 2017.
\newblock Adapting to learner errors with minimal supervision.
\newblock \emph{Computational Linguistics}, 43(4):723--760.

\bibitem[{Rychalska et~al.(2019)Rychalska, Basaj, Gosiewska, and
  Biecek}]{rychalska2019models}
Barbara Rychalska, Dominika Basaj, Alicja Gosiewska, and Przemys{\l}aw Biecek.
  2019.
\newblock Models in the wild: On corruption robustness of neural nlp systems.
\newblock In \emph{International Conference on Neural Information Processing},
  pages 235--247. Springer.

\bibitem[{{\v S}ebesta et~al.(2019){\v S}ebesta, Bed{\v r}ichov{\'a}, {\v
  S}ormov{\'a}, {\v S}tindlov{\'a}, Hrdli{\v c}ka, Hrdli{\v c}kov{\'a}, Hana,
  Petkevi{\v c}, Jel{\'{\i}}nek, {\v S}kodov{\'a}, Jane{\v s},
  Lund{\'a}kov{\'a}, Skoumalov{\'a}, Sl{\'a}dek, Pierscieniak, Toufarov{\'a},
  Straka, Rosen, N{\'a}plava, and Pol{\'a}{\v c}kov{\'a}}]{akces_gec}
Karel {\v S}ebesta, Zuzanna Bed{\v r}ichov{\'a}, Kate{\v r}ina {\v
  S}ormov{\'a}, Barbora {\v S}tindlov{\'a}, Milan Hrdli{\v c}ka, Tereza
  Hrdli{\v c}kov{\'a}, Ji{\v r}{\'{\i}} Hana, Vladim{\'{\i}}r Petkevi{\v c},
  Tom{\'a}{\v s} Jel{\'{\i}}nek, Svatava {\v S}kodov{\'a}, Petr Jane{\v s},
  Kate{\v r}ina Lund{\'a}kov{\'a}, Hana Skoumalov{\'a}, {\v S}imon Sl{\'a}dek,
  Piotr Pierscieniak, Dagmar Toufarov{\'a}, Milan Straka, Alexandr Rosen, Jakub
  N{\'a}plava, and Marie Pol{\'a}{\v c}kov{\'a}. 2019.
\newblock \href {http://hdl.handle.net/11234/1-3057} {{AKCES}-{GEC} grammatical
  error correction dataset for czech}.
\newblock {LINDAT}/{CLARIAH}-{CZ} digital library at the Institute of Formal
  and Applied Linguistics ({{\'U}FAL}), Faculty of Mathematics and Physics,
  Charles University.

\bibitem[{Sennrich et~al.(2016)Sennrich, Haddow, and
  Birch}]{sennrich2015improving}
Rico Sennrich, Barry Haddow, and Alexandra Birch. 2016.
\newblock Improving neural machine translation models with monolingual data.
\newblock In \emph{Proceedings of the 54th Annual Meeting of the Association
  for Computational Linguistics (Volume 1: Long Papers)}, pages 86--96.

\bibitem[{{\v{S}}ev{\v{c}}{\'i}kov{\'a}
  et~al.(2007){\v{S}}ev{\v{c}}{\'i}kov{\'a}, {\v{Z}}abokrtsk{\'y}, and
  Kr{\r{u}}za}]{sevcikova2007cnec}
Magda {\v{S}}ev{\v{c}}{\'i}kov{\'a}, Zden{\v{e}}k {\v{Z}}abokrtsk{\'y}, and
  Old{\v{r}}ich Kr{\r{u}}za. 2007.
\newblock Named entities in czech: Annotating data and developing ne tagger.
\newblock In \emph{Text, Speech and Dialogue}, pages 188--195, Berlin,
  Heidelberg. Springer Berlin Heidelberg.

\bibitem[{Straka et~al.(2019)Straka, Strakov\'{a}, and
  Haji\v{c}}]{straka2019evaluating}
Milan Straka, Jana Strakov\'{a}, and Jan Haji\v{c}. 2019.
\newblock \href {http://arxiv.org/abs/1908.07448} {Evaluating contextualized
  embeddings on 54 languages in {POS} tagging, lemmatization and dependency
  parsing}.

\bibitem[{Strakov{\'a} et~al.(2019)Strakov{\'a}, Straka, and
  Hajic}]{strakova-etal-2019-neural}
Jana Strakov{\'a}, Milan Straka, and Jan Hajic. 2019.
\newblock \href {https://doi.org/10.18653/v1/P19-1527} {Neural architectures
  for nested {NER} through linearization}.
\newblock In \emph{Proceedings of the 57th Annual Meeting of the Association
  for Computational Linguistics}, pages 5326--5331, Florence, Italy.
  Association for Computational Linguistics.

\bibitem[{Tjong Kim~Sang and
  De~Meulder(2003)}]{tjong-kim-sang-de-meulder-2003-introduction}
Erik~F. Tjong Kim~Sang and Fien De~Meulder. 2003.
\newblock \href {https://www.aclweb.org/anthology/W03-0419} {Introduction to
  the {C}o{NLL}-2003 shared task: Language-independent named entity
  recognition}.
\newblock In \emph{Proceedings of the Seventh Conference on Natural Language
  Learning at {HLT}-{NAACL} 2003}, pages 142--147.

\bibitem[{Vaswani et~al.(2017)Vaswani, Shazeer, Parmar, Uszkoreit, Jones,
  Gomez, Kaiser, and Polosukhin}]{vaswani2017attention}
Ashish Vaswani, Noam Shazeer, Niki Parmar, Jakob Uszkoreit, Llion Jones,
  Aidan~N Gomez, {\L}ukasz Kaiser, and Illia Polosukhin. 2017.
\newblock Attention is all you need.
\newblock In \emph{Advances in neural information processing systems}, pages
  5998--6008.

\bibitem[{Wang et~al.(2018)Wang, Singh, Michael, Hill, Levy, and
  Bowman}]{wang2018glue}
Alex Wang, Amanpreet Singh, Julian Michael, Felix Hill, Omer Levy, and Samuel
  Bowman. 2018.
\newblock Glue: A multi-task benchmark and analysis platform for natural
  language understanding.
\newblock In \emph{Proceedings of the 2018 EMNLP Workshop BlackboxNLP:
  Analyzing and Interpreting Neural Networks for NLP}, pages 353--355.

\bibitem[{Wolf et~al.(2019)Wolf, Debut, Sanh, Chaumond, Delangue, Moi, Cistac,
  Rault, Louf, Funtowicz, Davison, Shleifer, von Platen, Ma, Jernite, Plu, Xu,
  Scao, Gugger, Drame, Lhoest, and Rush}]{Wolf2019HuggingFacesTS}
Thomas Wolf, Lysandre Debut, Victor Sanh, Julien Chaumond, Clement Delangue,
  Anthony Moi, Pierric Cistac, Tim Rault, Rémi Louf, Morgan Funtowicz, Joe
  Davison, Sam Shleifer, Patrick von Platen, Clara Ma, Yacine Jernite, Julien
  Plu, Canwen Xu, Teven~Le Scao, Sylvain Gugger, Mariama Drame, Quentin Lhoest,
  and Alexander~M. Rush. 2019.
\newblock Huggingface's transformers: State-of-the-art natural language
  processing.
\newblock \emph{ArXiv}, abs/1910.03771.

\bibitem[{Xie et~al.(2018)Xie, Genthial, Xie, Ng, and
  Jurafsky}]{xie2018noising}
Ziang Xie, Guillaume Genthial, Stanley Xie, Andrew~Y Ng, and Dan Jurafsky.
  2018.
\newblock Noising and denoising natural language: Diverse backtranslation for
  grammar correction.
\newblock In \emph{Proceedings of the 2018 Conference of the North American
  Chapter of the Association for Computational Linguistics: Human Language
  Technologies, Volume 1 (Long Papers)}, pages 619--628.

\bibitem[{Yannakoudakis et~al.(2011)Yannakoudakis, Briscoe, and
  Medlock}]{yannakoudakis2011new}
Helen Yannakoudakis, Ted Briscoe, and Ben Medlock. 2011.
\newblock \href {https://www.aclweb.org/anthology/P11-1019} {A new dataset and
  method for automatically grading esol texts}.
\newblock In \emph{Proceedings of the 49th Annual Meeting of the Association
  for Computational Linguistics: Human Language Technologies-Volume 1}, pages
  180--189. Association for Computational Linguistics.

\bibitem[{Yannakoudakis et~al.(2018)Yannakoudakis, Øistein E~Andersen,
  Geranpayeh, Briscoe, and Nicholls}]{yannakoudakis2018}
Helen Yannakoudakis, Øistein E~Andersen, Ardeshir Geranpayeh, Ted Briscoe, and
  Diane Nicholls. 2018.
\newblock \href {https://doi.org/10.1080/08957347.2018.1464447} {Developing an
  automated writing placement system for esl learners}.
\newblock \emph{Applied Measurement in Education}, 31(3):251--267.

\bibitem[{Zeman et~al.(2018)Zeman, Haji{\v{c}}, Popel, Potthast, Straka,
  Ginter, Nivre, and Petrov}]{zeman-etal-2018-conll}
Daniel Zeman, Jan Haji{\v{c}}, Martin Popel, Martin Potthast, Milan Straka,
  Filip Ginter, Joakim Nivre, and Slav Petrov. 2018.
\newblock \href {https://doi.org/10.18653/v1/K18-2001} {{C}o{NLL} 2018 shared
  task: Multilingual parsing from raw text to universal dependencies}.
\newblock In \emph{Proceedings of the {C}o{NLL} 2018 Shared Task: Multilingual
  Parsing from Raw Text to Universal Dependencies}, pages 1--21, Brussels,
  Belgium. Association for Computational Linguistics.

\end{thebibliography}

\clearpage

\title{\titletext\\\large Supplementary Material}

\maketitle
\beginsupplement

\section{Morpho-syntactic Analysis}
\label{supp:sec:udpipe}

\begin{figure}[t!]
    \centering
    \includegraphics[width=.99\hsize]{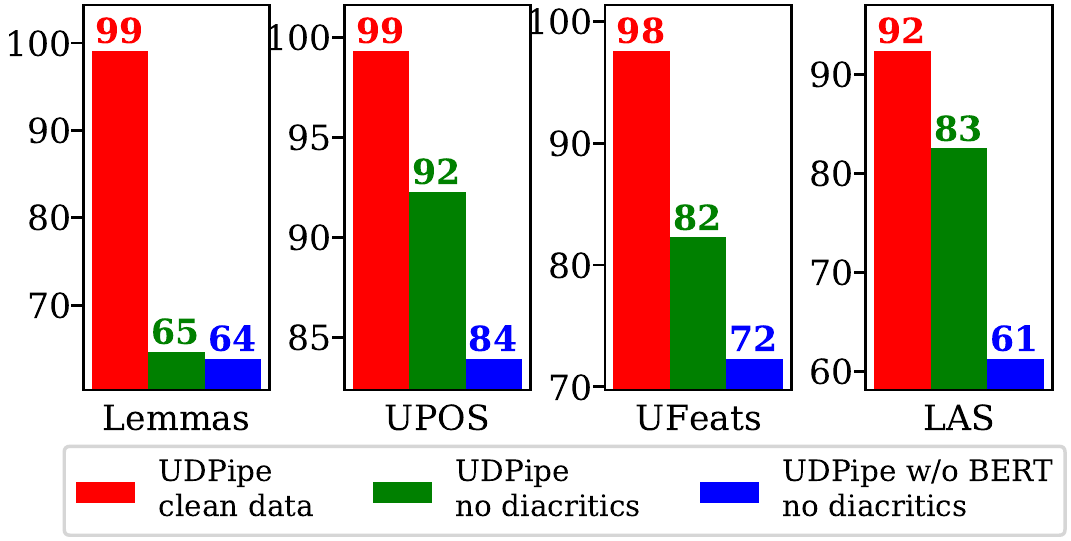}
    \caption{Morpho-syntactic tasks evaluated on Czech with diacritics (red) and without diacritics: using UDPipe with BERT-uncased (green) and UDPipe without BERT (blue).}
    \label{supp:fig:udpipe_diacritics}
\end{figure}

Figure~\ref{supp:fig:udpipe_diacritics} shows the importance of diacritics handling in all morpho-syntactic tasks in Czech, especially for lemmatization. We initially expected that the diacritics-ignoring BERT model used in the UDPipe architecture would make the system resilient to missing diacritics. The performance is indeed superior to a UDPipe variant not using any BERT model, but is still severely reduced on texts without diacritics. We attribute the decrease by the fact, that apart from using BERT embeddings UDPipe utilizes its own word and character-level word embeddings.

\begin{figure}[t!]
    \centering
    \includegraphics[width=.99\hsize]{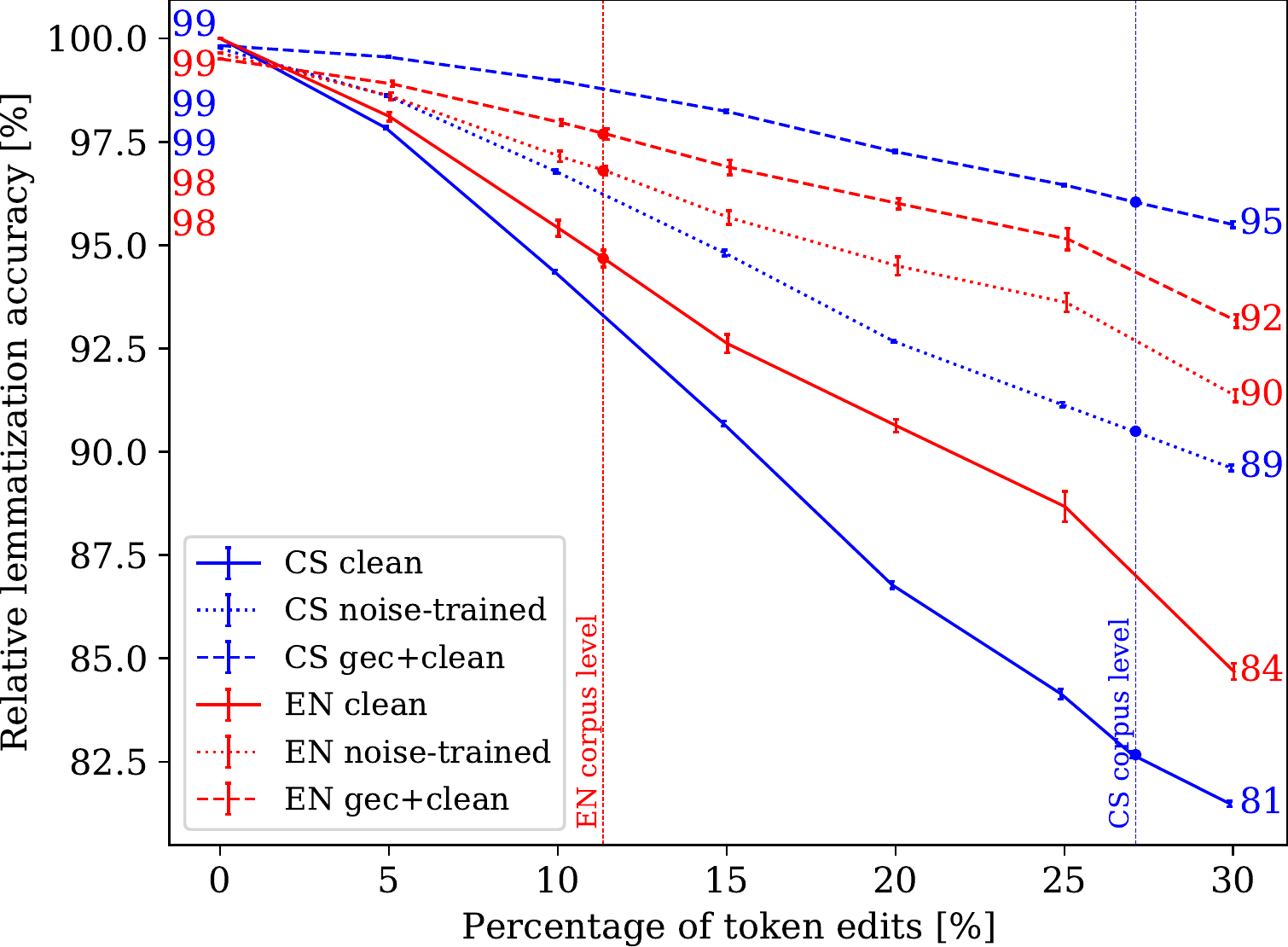}
    \caption{Relative lemmatization accuracy on increasing percentage of token edits based on Second Learners profile, using clean model, noise-trained model and grammar-error-correction.}
    \label{supp:fig:udpipe_corr_lemmas}
\end{figure}

Figure~\ref{supp:fig:udpipe_corr_lemmas} presents lemmatizer robustness with respect to increasing percentage of token errors in the text. Lines marked as \textit{CS clean} and \textit{EN clean} display the performance of the clean models on texts with up to $30\%$ of token edits. Additionally, we present the performance of our two noise-mitigating strategies: training model on noisy texts (\textit{noise-trained}) and preprocessing with an external language corrector (\textit{gec+clean}).

\section{Named Entity Recognition}
\label{supp:sec:ner}

\begin{figure}[t!]
    \centering
    \includegraphics[width=.99\hsize]{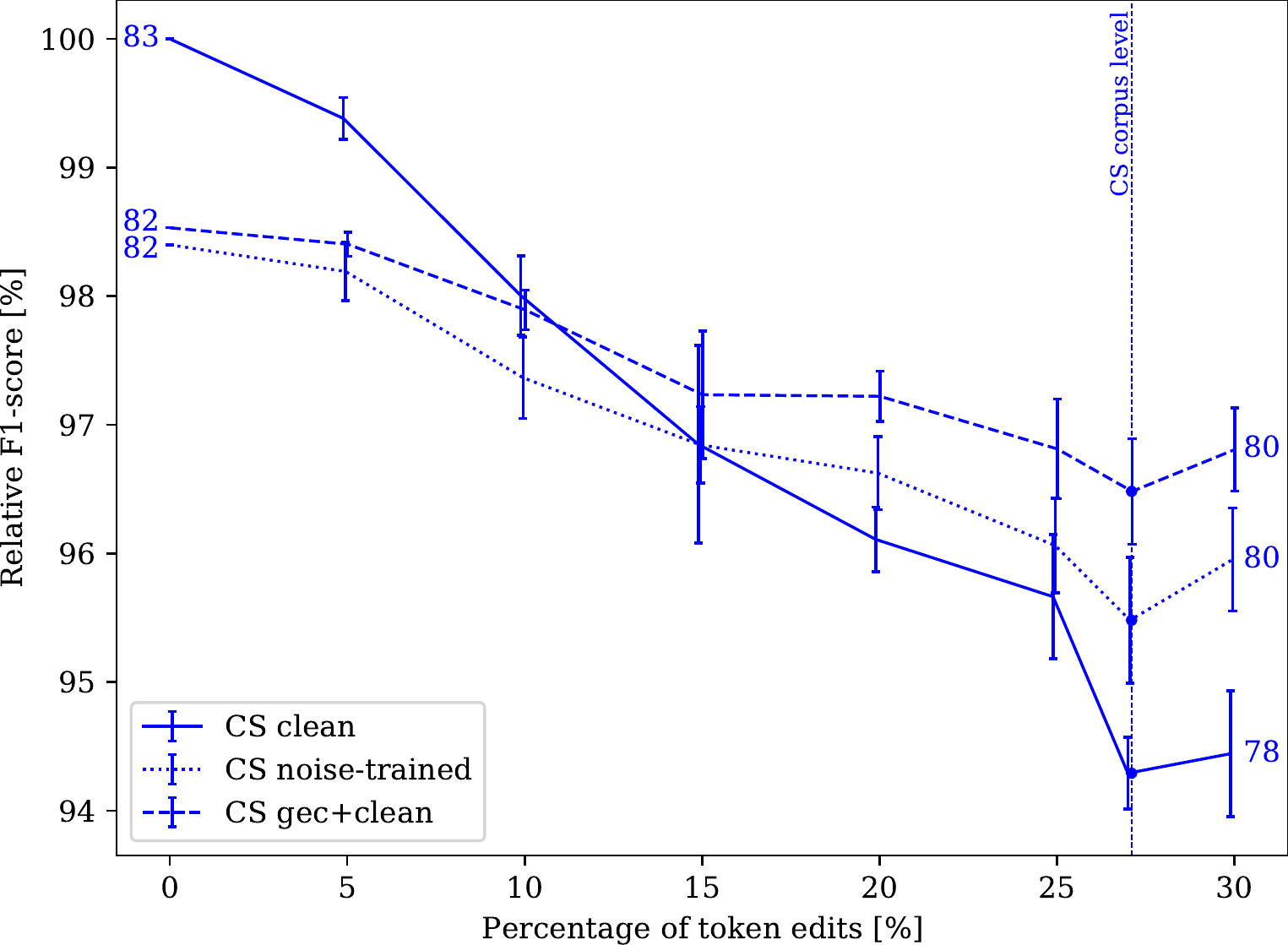}
    \includegraphics[width=.99\hsize]{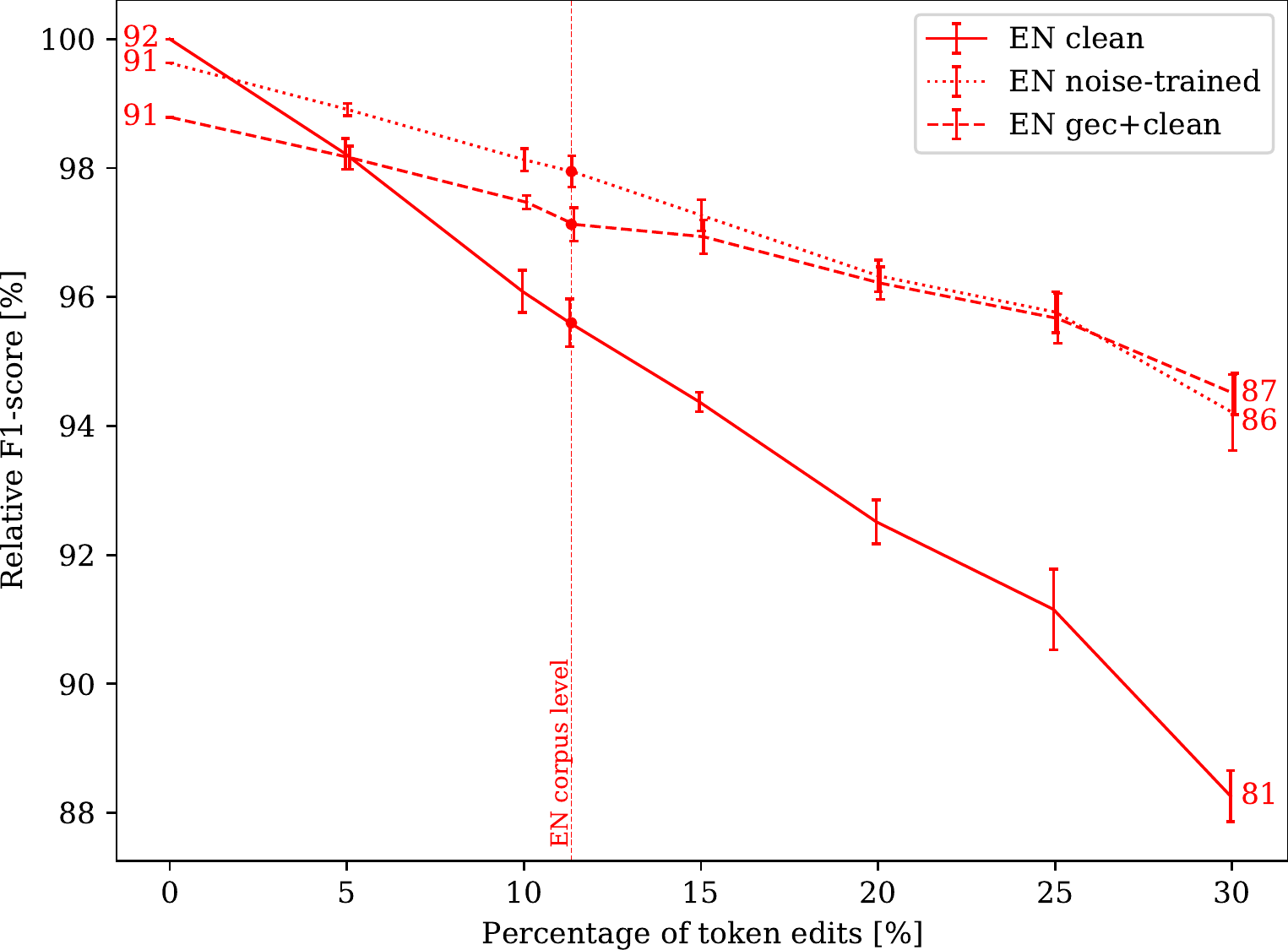}
    \caption{NER F1 relative to original data F1 on increasing percentage of token edits, using clean model, noise-trained model and grammatical-error-correction. Numbers near lines are absolute F1 values. \textbf{Upper} Czech second learners profile. \textbf{Lower} English second learners profile.}
    \label{supp:fig:ner_csen}
\end{figure}

The impact of increasing percentage of token edits on NER in Czech and
English second learning profiles is presented in Figure~\ref{supp:fig:ner_csen},
comparing a clean model, a model trained on noised inputs and a model paired with GEC.

We further document the casing effect on the NER task performance. As we can see in Table~\ref{supp:table:ner_all_lower}, the performance of all models drops dramatically when the text is completely lowercased. The biggest drops are on English and Czech, while the difference on German is not that substantial. We address this to the fact that while the prevalence amount of words with first uppercased letter in English and Czech are named entities, it does not hold in German.

\begin{table}[t!]
    \begin{center}
        \setlength{\tabcolsep}{13pt}
        \begin{tabular}{lcc}\toprule
            language    & original  & uncased   \\\midrule
            English     & 91.68     & 49.01     \\
            German      & 82.65     & 73.61     \\
            Czech       & 83.07     & 50.69     \\\bottomrule
        \end{tabular}
    \end{center}
    \caption{NER F1 on original and uncased text.}
    \label{supp:table:ner_all_lower}
\end{table}

\vfill\eject
\leavevmode
\vfill\eject
\leavevmode







\end{document}